\documentclass{article} 
\usepackage[final]{colm2026_conference}

\usepackage{microtype}
\usepackage{subcaption}
\usepackage{adjustbox}
\usepackage{hyperref}
\usepackage{graphicx}
\usepackage{url}
\usepackage{booktabs}
\usepackage{algorithm}
\usepackage{algpseudocode}
\usepackage{tcolorbox}
\usepackage{enumitem}
\usepackage{caption}
\usepackage{placeins}
\usepackage{wrapfig}
\tcbuselibrary{breakable, skins}

\usepackage{amsmath,amsfonts,bm}

\def\eqref#1{equation~\ref{#1}}

\def\1{\bm{1}}

\DeclareMathAlphabet{\mathsfit}{\encodingdefault}{\sfdefault}{m}{sl}
\SetMathAlphabet{\mathsfit}{bold}{\encodingdefault}{\sfdefault}{bx}{n}

\usepackage{float}

\newtcolorbox{promptbox}[1][]{
  enhanced,
  breakable,
  colback=orange!3,
  colframe=orange!70!black,
  coltitle=white,
  fonttitle=\bfseries\small,
  title=#1,
  boxrule=0.5pt,
  arc=1pt,
  left=8pt,
  right=8pt,
  top=6pt,
  bottom=6pt,
  fontupper=\small
}

\usepackage{lineno}

\definecolor{darkblue}{rgb}{0, 0, 0.5}
\hypersetup{colorlinks=true, citecolor=darkblue, linkcolor=darkblue, urlcolor=darkblue}

\title{Filtered Reasoning Score: Evaluating Reasoning Quality on a Model's Most Confident Traces}

\author{Manas Pathak, Xingyao Chen, Shuozhe Li, Amy Zhang, Liu Leqi \\
University of Texas at Austin \\
\texttt{\{manaspathak, caspar.cxy06, shuozhe.li, leqiliu\}@utexas.edu}}

\begin{document}
\raggedbottom

\ifcolmsubmission
\linenumbers  
\fi

\maketitle

\begin{abstract}

Should we trust Large Language Models (LLMs) with high accuracy?
LLMs achieve high accuracy on reasoning benchmarks, but correctness alone does not reveal the quality of the reasoning used to produce it. This highlights a fundamental limitation of outcome-based evaluation: models may arrive at correct answers through flawed reasoning, and models with substantially different reasoning capabilities can nevertheless exhibit similar benchmark accuracy, for example due to memorization or over-optimization. 
In this paper, we ask: given existing benchmarks, can we move beyond outcome-based evaluation to assess the quality of reasoning itself? We seek metrics that (1) differentiate models with similar accuracy and (2) are robust to variations in input prompts and generation configurations. To this end, we propose a reasoning score that evaluates reasoning traces along dimensions such as faithfulness, coherence, utility, and factuality.
A remaining question is how to aggregate this score across multiple sampled traces. Naively averaging them is undesirable, particularly in long-horizon settings, where the number of possible trajectories grows rapidly, and low-confidence correct traces are more likely to be coincidental. To address this, we introduce the Filtered Reasoning Score (FRS), which computes reasoning quality using only the top-$K\%$ most confident traces. 
Evaluating with FRS, models that are indistinguishable under standard accuracy exhibit significant differences in reasoning quality.
Moreover, models with higher FRS on one benchmark tend to perform better on other reasoning benchmarks, in both accuracy and reasoning quality. Together, these findings suggest that FRS complements accuracy by capturing a model's transferable reasoning capabilities.
We open source our evaluation codebase: \url{https://github.com/HumainLab/filtered_reasoning_score_evaluation}.

  \end{abstract}

\begin{figure}[h]
  \centering
  \begin{tcolorbox}[enhanced, 
    colback=white, 
    colframe=black!60, 
    boxrule=0.6pt, 
    arc=2pt,
    left=5pt, right=5pt, top=2pt, bottom=2pt,
    fontupper=\footnotesize,
    title={\footnotesize\bfseries Two correct answers, very different reasoning quality},
    coltitle=white, 
    colbacktitle=black!70]
    
  \textbf{Trace A} \hfill \colorbox{green!15}{\textit{Score: 100/100}}\\[1pt]
  \textit{Problem:} Octagon with same perimeter as hexagon of side 16\,cm. Find octagon side length.\\
  Perimeter $= 6 \times 16 = 96$\,cm. Octagon has 8 sides: $8s = 96 \Rightarrow s = 12$.\quad $\boxed{12}$
  \tcbline
  \textbf{Trace B} \hfill \colorbox{red!15}{\textit{Score: 25/100}}\\[1pt]
  \textit{Problem:} GCF of 6432 and 132, increased by 11.\\
  Lists factors, concludes GCF $= 4$, gets $4{+}11{=}15$ \textit{[\textcolor{red}{wrong}]} $\rightarrow$ ... 
  \quad $\boxed{23}$
  
  \end{tcolorbox}
  \vspace{-16pt}
  \caption{Two traces from different models produce correct final answers and receive the same Pass@1 score, yet their reasoning quality is vastly different.}
  \label{box:two-traces}
\end{figure}

\section{Introduction}

Large language models have advanced rapidly in recent years, 
with newer models achieving higher scores on an expanding set of reasoning benchmarks~\citep{hendrycks2021math, cobbe2021gsm8k, rein2024gpqa}. 
Yet the way we evaluate these models has not kept pace. 
The dominant paradigm remains final-answer accuracy: a model is scored by how often it produces the correct output, with no regard for the reasoning process that produced it.

This paradigm is increasingly inadequate. 
Models can produce flawed reasoning that still leads to correct answers~\citep{lightman2023lets, uesato2022solving, turpin2023language}. 
As a result, accuracy gains do not reliably reflect improvements in reasoning quality~\citep{xia2024evaluating}, and benchmark saturation further reduces their ability to distinguish between models~\citep{deveci2025ouroboros}. 
Moreover, outcome-based evaluation can be sensitive to prompt choice and generation configuration, further obscuring differences in underlying reasoning ability \citep{hochlehnert2025sober}.

These limitations motivate a shift from evaluating outcomes to evaluating reasoning itself. In this work, we ask: \textit{can we use existing benchmarks to assess the quality of reasoning, beyond final-answer correctness?} We seek metrics that (1) differentiate models with similar accuracy with respect to their underlying reasoning capabilities, and (2) are robust to variations in prompts and generation configurations. To this end, we propose to evaluate reasoning traces directly using a rubric-based reasoning score that measures properties such as faithfulness, coherence, utility, and factuality (Section \ref{sec:rs}).

A second challenge is aggregation: models can produce many reasoning traces for the same problem, with substantial variation in both likelihood and quality. While averaging over all sampled traces may provide an unbiased estimate of its expected reasoning score, this is not the quantity most relevant in practice. Deployed systems typically act on one selected output, often chosen explicitly or implicitly based on confidence \citep{kang2025scalable,wen2025abstention, geifman2017selective}. 
Evaluating all traces equally therefore gives disproportionate influence to low-probability trajectories that are unlikely to be selected in practice.

To address this, we introduce the \textbf{Filtered Reasoning Score (FRS)}, which evaluates reasoning quality on the most confident subset of traces in terms of their faithfulness, coherence, utility, and factuality. For each problem, we sample multiple reasoning traces, estimate per-trace confidence from token-level probabilities, and retain only the top-$K\%$ by confidence before computing the final score. This design focuses evaluation on the high-confidence region of the model’s reasoning trace distribution, where outputs are more representative of the model’s intended inference behavior. A high FRS therefore requires both strong reasoning and high confidence on high-quality reasoning traces (Section \ref{sec:frs_def}).

\begin{figure}[htbp]
    \centering
    \makebox[\linewidth][c]{%
        \includegraphics[width=1.2\linewidth]{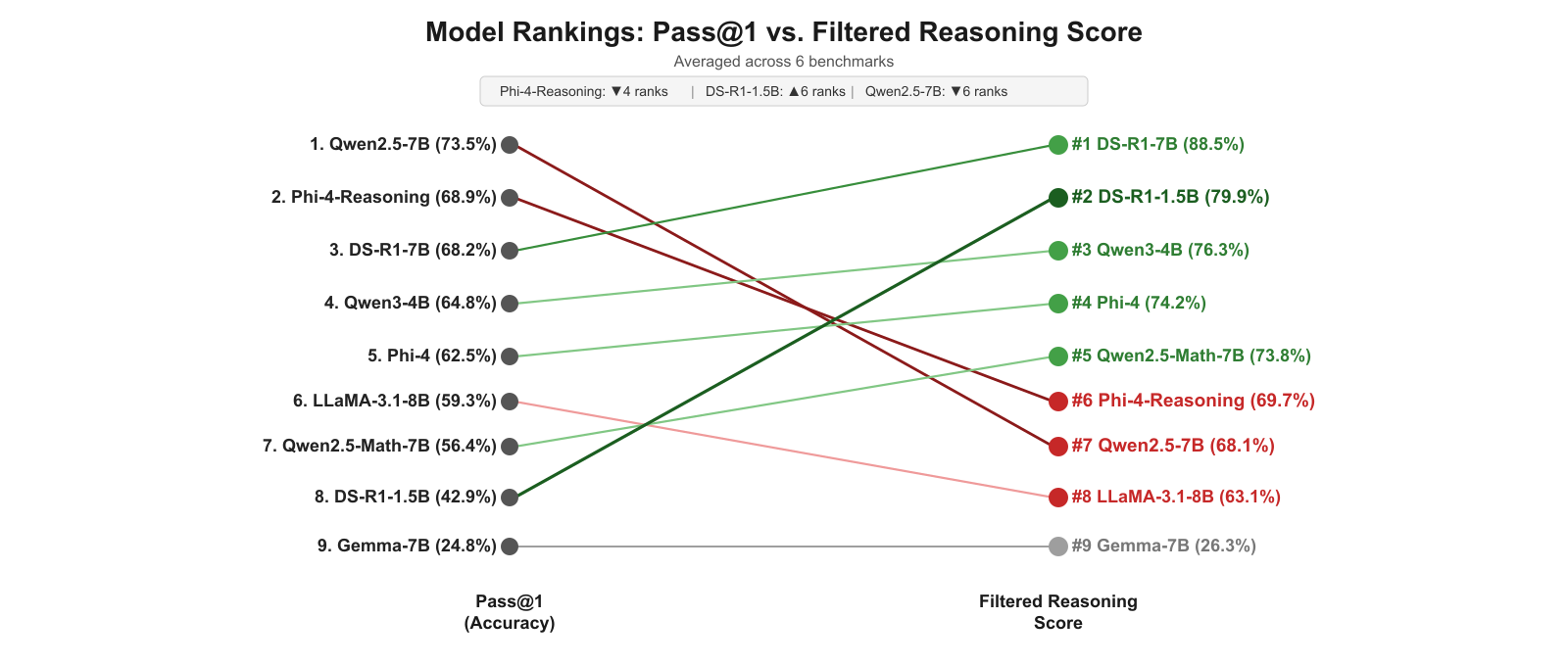}}
    \vspace{-14pt}
    \caption{Model rankings under greedy pass@1 (left) vs.\ Filtered Reasoning Score with $K = 10\%$ (right), averaged across 6 benchmarks. Green lines indicate models that rise under FRS; red lines indicate models that fall. Qwen2.5-7B: $\#1 \to \#7$; DS-R1-1.5B: $\#8 \to \#2$.}
    \label{fig:ranking_reversals}
\end{figure}
\vspace{-6pt}

Empirically, FRS complements accuracy by identifying distinguishing characteristics among models with similar accuracy. 
We evaluate FRS across 9 open-weight models (1.5B to 14B parameters) and 6 reasoning benchmarks (Figure~\ref{fig:ranking_reversals}). On MATH500~\citep{lightman2023lets}, two models with identical greedy accuracy (63.6\%) differ by 16.5 FRS points. The highest-accuracy model overall drops from \#1 to \#7 under FRS, while a model ranked \#8 by accuracy rises to \#2. Across benchmarks, models within $0.7\%$ accuracy differ by up to 18.8 FRS points. These differences reflect systematic variation in how models assign confidence over reasoning traces: some models confidently produce flawed reasoning that may lead to correct answers, while others more reliably assign high confidence to well-reasoned solutions. Such distinctions are invisible under accuracy-based evaluation, yet directly impact which reasoning is generated in practice.

Our contributions are:
\begin{enumerate}[nosep, leftmargin=1.5em]
    \item \textbf{We identify confidence-conditioned reasoning quality as a distinct evaluation target beyond answer accuracy.} Correctness alone does not capture reasoning quality: two traces can reach the same answer while differing substantially in faithfulness, coherence, utility, and factuality. This matters especially when LLM systems prioritize high-confidence outputs, making it important to evaluate whether a model’s most confident reasoning traces are of high quality.
    
    \item \textbf{We introduce the Filtered Reasoning Score (FRS).} FRS evaluates reasoning quality on the most confident traces, requiring both strong reasoning and high confidence on well-reasoned solutions.
    
    \item \textbf{We show that FRS reveals structure hidden by accuracy-based evaluation.} Across 9 open-weight models and 6 reasoning benchmarks, FRS exposes ranking reversals, large separations among accuracy-similar models, and systematic differences in how confidence relates to reasoning quality, with all key results supported by non-overlapping bootstrap 95\% CIs. FRS is also the only metric among six candidates that significantly predicts whether confidence-based selection improves or degrades reasoning quality ($r{=}0.49$, $p{<}0.001$; Appendix~\ref{app:selection_gain}).
\end{enumerate}

\section{Related Work}

\paragraph{Reasoning evaluation and benchmark saturation.} Most reasoning benchmarks evaluate final-answer correctness~\citep{cobbe2021gsm8k, hendrycks2021math, rein2024gpqa}, but accuracy improvements do not reliably track reasoning quality~\citep{xia2024evaluating, mondorf2024beyond}. Process-level evaluation has received growing attention through process reward models (PRMs)~\citep{lightman2023lets, uesato2022solving} and rubric-based taxonomies~\citep{lee2025evaluating}; \citep{turpin2023language} show that chain-of-thought explanations \citep{wei2022chain} can be unfaithful to the model's actual reasoning \citep{li2026care, li2026learning}. PRMs are the closest existing alternative to our rubric-based scorer; we compare against a PRM head-to-head in Appendix~\ref{app:prm} and find that the two are uncorrelated over the full trace distribution ($\rho = 0.15$) but converge once traces are confidence-filtered ($\rho = 0.78$), indicating that the confidence filter, not the scorer, is the source of FRS's predictive signal. Meanwhile, benchmark saturation is eroding the discriminative power of existing evaluations~\citep{deveci2025ouroboros, hashemi2025emdm}. Our approach addresses saturation differently: rather than creating harder problems, we condition evaluation on a model's own confidence signal, exposing structure invisible to any accuracy threshold.

\paragraph{Confidence estimation and calibration.} LLM confidence estimation spans verbalized methods~\citep{xiong2024can, tian2023just}, sampling-based approaches such as self-consistency~\citep{wang2022selfconsistency} and semantic entropy~\citep{kuhn2023semantic, farquhar2024semantic}, and logit-based methods~\citep{hendrycks2017baseline, kadavath2022language}; \citet{geng2024survey} provide a comprehensive survey. Recent work on targeted token aggregation~\citep{kang2025selfcertainty, fu2025deepconf, devic2025trace} informs our confidence estimator design (Section~\ref{sec:confidence_estimator}). On calibration, \citet{zeng2025thinkingoutloud} find that RL-trained models achieve better confidence calibration, and \citet{knowwhenyourewrong2026} show that post-RL SFT can restore calibration degraded by GRPO training. FRS is complementary: calibration asks whether confidence tracks \emph{correctness}; FRS asks whether high confidence selects for strong \emph{reasoning}, a question calibration cannot answer, since models can reach correct answers through flawed reasoning. Selective prediction~\citep{elyaniv2010selective, geifman2017selective, geifman2019selectivenet, wen2025abstention} studies when to abstain but does not evaluate the quality of accepted traces; FRS fills this gap.

\paragraph{LLM-as-a-judge.} We use GPT-4o-mini as a rubric-based judge following the paradigm established by \citet{zheng2023judging} and extended by \citet{liu2023geval}. We validate against GPT-4o, Claude Sonnet 4.5, and human annotators (Appendix~\ref{app:judge}).
\section{Methodology}
\label{sec:methodology}

\subsection{Reasoning Quality Evaluation}\label{sec:rs}

To capture the quality of a model's reasoning process, we score chain-of-thought (CoT) traces along four dimensions following the taxonomy of~\citet{lee2025evaluating}: \textbf{faithfulness} (internal consistency without hidden shortcuts), \textbf{coherence} (logical flow between steps), \textbf{utility} (whether each step contributes to the solution with correct calculations), and \textbf{factuality} (grounding in the problem context without hallucinations). We provide the full scoring rubric in Appendix~\ref{app:rubric}. Empirically, the four dimensions organize into two axes: faithfulness, coherence, and utility are mutually correlated (Spearman $\rho \in [0.90, 0.98]$) and capture a dominant logical-quality axis, while factuality is largely independent ($\rho \in [0.35, 0.61]$ with the other three) and contributes a distinct grounding signal. A leave-one-dimension-out ablation shows FRS is stable to removing any single logical-axis dimension, and we retain all four following~\citet{lee2025evaluating}; see Appendix~\ref{app:dim_ablation} for the full analysis.

We use GPT-4o-mini as an automated evaluator following the LLM-as-a-judge paradigm~\citep{zheng2023judging}. The judge receives the question, the model's full response (CoT trace and final answer), the ground-truth answer, and the scoring rubric, and produces a score from 1 (very poor) to 5 (excellent) for each dimension. We validate judge reliability in three ways: cross-model agreement with GPT-4o (82\% within $\pm$1) and Claude Sonnet 4.5 (75\% within $\pm$1), and a blind human annotation study with 10 annotators and 500 labeled samples, achieving 78\% within $\pm$1 agreement and a Spearman correlation of $\rho = 0.73$ (Appendix~\ref{app:judge}). The \textbf{Reasoning Score} is the average across the four dimensions, normalized to $[0, 1]$:
\begin{equation}
    \text{ReasoningScore}(r) = \frac{1}{4} \sum_{d \in \{\text{faith., coh., util., fact.}\}} \frac{s_d(r) - 1}{4}
\end{equation}
where $s_d(r) \in \{1, 2, 3, 4, 5\}$ is the judge's score for dimension $d$ on reasoning trace $r$.

\subsection{Per-Trace Confidence Estimation}
\label{sec:confidence_estimator}

The first stage of FRS is to assign each reasoning trace a scalar confidence score that can be computed without correctness labels. We use a logit-based estimator derived from token-level probabilities. Rather than averaging over the full trace, we focus on its low-probability tail, since these tokens concentrate the model's uncertainty, whereas many high-probability tokens are generic and contribute little to distinguishing strong traces from weak ones.

Concretely, for each input $x_i$, the model generates a reasoning trace
$r_i = (t_1, \ldots, t_{L_i})$. For each token, we extract its conditional probability
$P(t_j) = P(t_j \mid x_i, t_1, \ldots, t_{j-1})$. We then define trace-level confidence as
\begin{equation}
    C(r_i) = \frac{1}{|\mathcal{T}_{\text{low}}|} 
    \sum_{t \in \mathcal{T}_{\text{low}}} P(t),
\end{equation}
where $\mathcal{T}_{\text{low}}$ is the set of tokens in $r_i$ with probabilities below the $p$th percentile.

We use $p{=}10\%$ as the default high-confidence operating point, targeting the region where low-probability tokens most concentrate the model's uncertainty. This value is chosen once via a coarse calibration sweep and is stable across nearby choices in the 5--15\% range (Appendix~\ref{app:lowprob}). We note that this estimator is empirically rather than theoretically motivated, and none of the paper's conclusions are load-bearing on it: rankings are qualitatively consistent across $p \in \{10, 20, 30, 40, 50\}$ and across alternative confidence estimators, including full-trace mean log-probability ($\rho = 0.983$) and self-consistency confidence computed purely from answer agreement without any logit access ($\rho = 0.80$; Appendices~\ref{app:lowprob},~\ref{app:confidence_proxies}). These confidence scores define the ranking used in the next stage of FRS, where we retain only the most-confident traces for reasoning evaluation.

\subsection{Filtered Reasoning Score}
\label{sec:frs_def}

The second stage of FRS uses the confidence scores above to construct a filtered set of traces, and then measures reasoning quality only on that set. For each problem $x_i$, we sample $k{=}16$ reasoning traces at temperature $T{=}0.7$ using chain-of-thought prompting, compute $C(r_i^{(j)})$ for each trace, and pool all traces within a model--benchmark pair. We then rank these traces by confidence and retain the top $K\%$ as the \emph{filtered set}.

We use $K{=}10$ as the default, targeting the region where a model's outputs would be trusted most in deployment; rankings are qualitatively consistent across $K \in \{10, 20, 30, 40, 50\}$ (Appendix~\ref{app:fullresults}). Because traces are pooled globally before ranking, one might worry that the filtered set concentrates on easy problems where the model is uniformly confident. We verify this is not the case: more than 64\% of problems contribute at least one trace to the top-10\% set, and the set draws \emph{more} from hard problems (22.3\% of selected traces come from problems solved at most 4/16 times) than from uniformly easy ones (16.4\% from problems solved 16/16); see Appendix~\ref{app:selected_composition}.

To estimate FRS efficiently, we partition the top 50\% of traces into five disjoint confidence bins (0--10\%, 10--20\%, \ldots, 40--50\%) and randomly sample 50 traces per bin for each model--benchmark pair. These sampled traces are scored by GPT-4o-mini using the rubric in Section~3.1. The Filtered Reasoning Score at threshold $K$ is then
\begin{equation}
    \text{FRS}_{K} = \frac{1}{|\mathcal{S}_K|} \sum_{r \in \mathcal{S}_K} \text{ReasoningScore}(r),
\end{equation}
where $\mathcal{S}_K$ denotes all sampled traces from bins within the top $K\%$.

This construction makes the two components of FRS explicit: it rewards models that produce strong reasoning, but only when that reasoning appears in the traces that their own confidence ranks highest. A model with strong reasoning but misaligned confidence or weak reasoning overall will therefore score poorly.
\section{Experiments}

\subsection{Setup}
\label{sec:setup}

We evaluate 9 open-weight models spanning 1.5B to 14B parameters: DeepSeek-R1-Distill-Qwen-1.5B and -7B~\citep{deepseek2025r1}, LLaMA-3.1-8B-Instruct~\citep{dubey2024llama3}, Qwen2.5-7B-Instruct and Qwen2.5-Math-7B~\citep{yang2024qwen25}, Gemma-7B~\citep{team2024gemma}, Phi-4 and Phi-4-Reasoning~\citep{abdin2024phi4}, and Qwen3-4B in thinking mode~\citep{qwen2025qwen3}, covering general-purpose, math-specialized, and reasoning-optimized architectures. We evaluate on GSM8K~\citep{cobbe2021gsm8k}, MATH500~\citep{hendrycks2021math, lightman2023lets}, SVAMP~\citep{patel2021svamp}, AQuA~\citep{ling2017aqua}, GPQA~\citep{rein2024gpqa}, and CommonsenseQA~\citep{talmor2019commonsenseqa}, spanning arithmetic, mathematical, commonsense, and graduate-level science reasoning. For each problem, we sample $k{=}16$ reasoning traces at $T{=}0.7$; confidence and FRS are computed as described in Sections~\ref{sec:confidence_estimator}--\ref{sec:frs_def}.

\begin{wrapfigure}{R}{0.48\columnwidth}
    \vspace{-14pt}
    \centering
    \includegraphics[width=0.46\columnwidth]{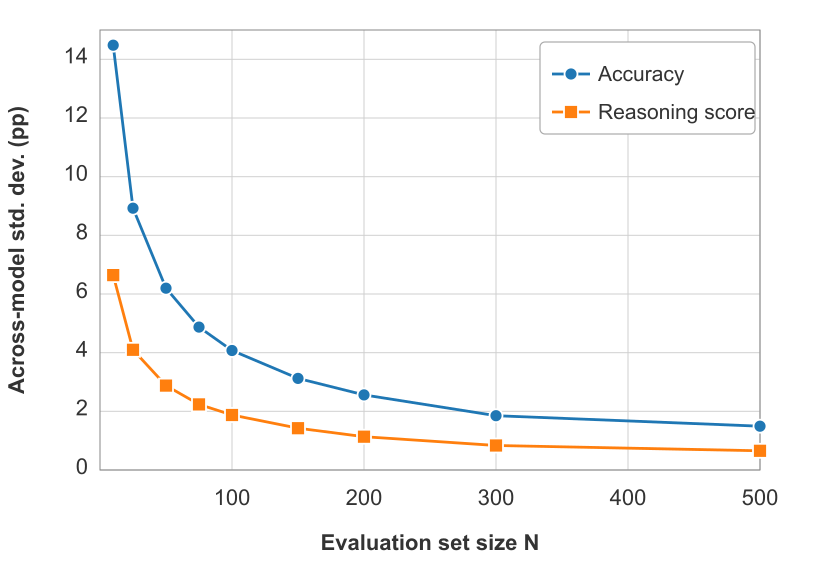}
    \vspace{-6pt}
    \caption{Median across-model std.\ dev.\ vs.\ evaluation set size $N$ (54 model-dataset combinations). The reasoning score has lower variance for all sample sizes. Per-benchmark breakdowns in Appendix~\ref{app:convergence}.}
    \label{fig:convergence}
    \vspace{-20pt}
\end{wrapfigure}

\subsection{Why Reasoning Quality, Not Just Accuracy}
\label{sec:reasoning_stable}

Before conditioning on confidence, we establish that reasoning quality adds information beyond accuracy. We show that the reasoning score is more stable and converges faster than accuracy.

\paragraph{Reasoning score produces more stable rankings.} We measure three stability metrics (average pairwise Spearman $\rho$, percentage of rank inversions, and average maximum rank swing) across four evaluation conditions (direct prompting, few-shot, CoT with $T{=}0$, CoT with $T{=}0.7$). On 5 of 6 benchmarks, reasoning score produces higher rank correlations, fewer inversions, and smaller rank swings than accuracy. Averaged across benchmarks, reasoning-based rankings achieve Spearman $\rho$ of 0.72 vs.\ 0.49 for accuracy, with 33\% fewer inversions (19.9\% vs.\ 29.9\%) and 31\% smaller maximum rank swings (2.4 vs.\ 3.5 positions). Full results are in Table~\ref{tab:ranking_stability} (Appendix~\ref{app:ranking_stability}).

\paragraph{Reasoning score converges faster.} Reasoning scores reach stable rankings with fewer evaluation problems than accuracy (Figure~\ref{fig:convergence}). Across 54 model-dataset combinations, the median standard deviation of reasoning scores is lower than that of accuracy at every sample size, and the gap is largest at small $N$: at $N{=}25$, accuracy fluctuates at roughly 9pp while reasoning score is already below 4pp. However, unfiltered reasoning score is insufficient: models with similar averages can have opposite confidence-quality relationships. For example, two models may both average 85 on reasoning quality, yet one improves to 97 in its high-confidence region while the other drops to 65. Exposing this structure requires conditioning on confidence, which is what FRS does.


\subsection{Why Confidence-Based Filtering Is Necessary}
\label{sec:confidence}

Having established that reasoning quality is a richer signal than accuracy, we address a key aggregation challenge: how should reasoning quality be summarized across multiple sampled traces? Averaging equally over all traces is not the quantity most relevant in practice. Deployed systems typically act on one selected output, often chosen explicitly or implicitly based on confidence~\citep{kang2025scalable, wen2025abstention, geifman2017selective}, giving disproportionate influence to low-probability trajectories unlikely to be surfaced in deployment. We therefore ask: \emph{is better reasoning concentrated in the model's high-confidence region?}

\begin{figure}[h!]
    \centering
    \includegraphics[width=0.72\linewidth]{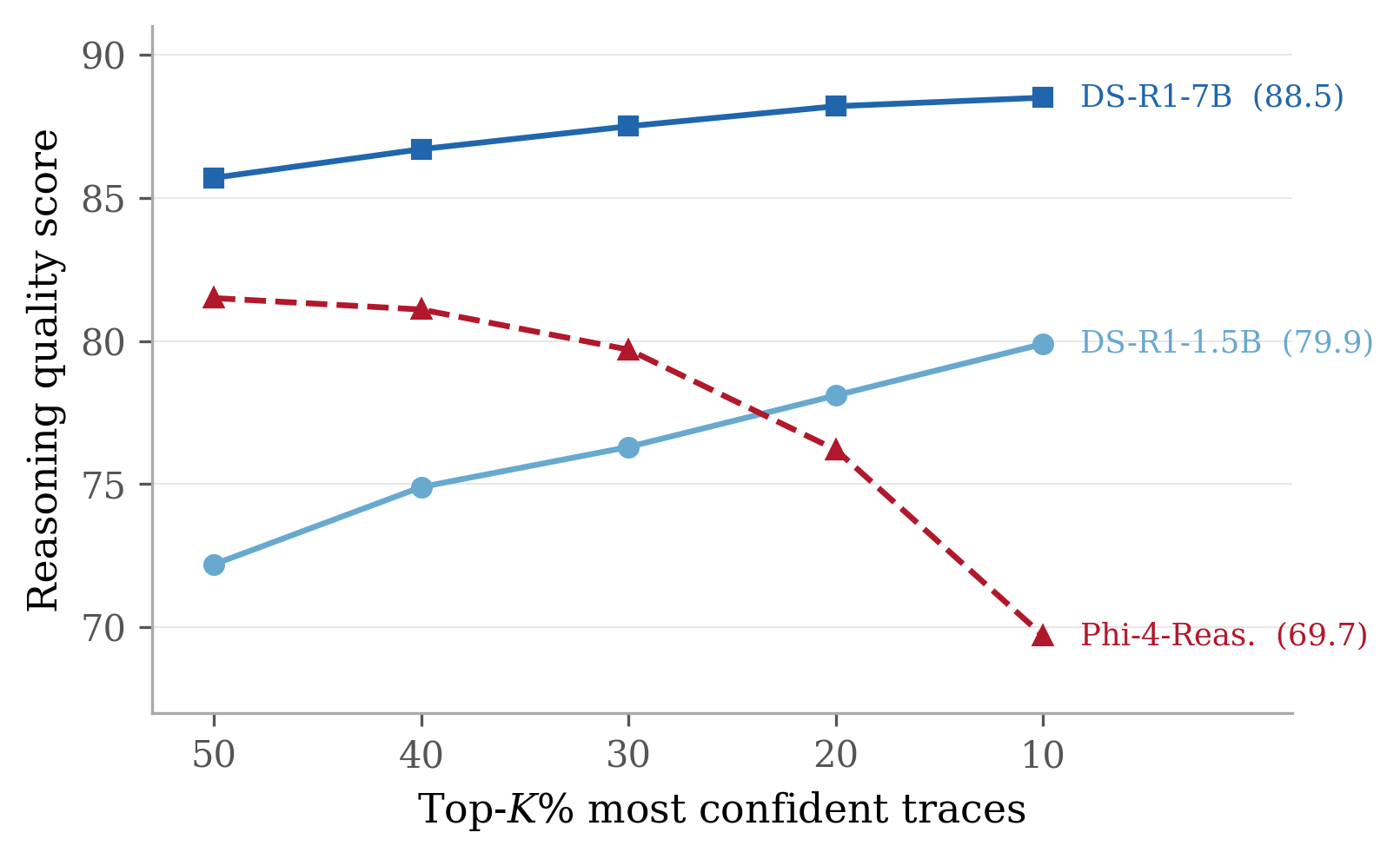}
    \caption{Reasoning quality on the top $K\%$ most-confident traces, averaged across 6 benchmarks, for three representative models. As the confidence filter tightens (lower $K$), quality improves for some models and degrades for others. Full results for all 9 models are in Appendix~\ref{app:fullresults}.}
    \label{fig:topk_representative}
\end{figure}

Figure~\ref{fig:topk_representative} plots average reasoning quality at each threshold $K \in \{50,40,30,20,10\}$ for three representative models. For DS-R1-7B, restricting to higher-confidence traces yields markedly better reasoning: its score rises from 85.7 at $K{=}50\%$ to 88.5 at $K{=}10\%$. DS-R1-1.5B shows an even steeper gain, climbing from 72.2 to 79.9. By contrast, Phi-4-Reasoning moves in the opposite direction, dropping from 81.5 to 69.7, meaning its most-confident traces contain \emph{worse} reasoning than its average output.

Two models with the same average reasoning score can therefore differ sharply in the quality their confidence selects. Unfiltered averaging cannot distinguish a model whose confidence surfaces its best reasoning from one that surfaces its worst. We evaluate reasoning quality only in the high-confidence region, where deployed systems actually operate. The next section fixes $K{=}10\%$ and reports FRS across all models and benchmarks.

\section{FRS Results: What Confidence Actually Selects}
\label{sec:frs_results}

We now evaluate reasoning quality at $K{=}10\%$ across all 9 models and 6 benchmarks. Table~\ref{tab:frs_main} reports the full results, and Figure~\ref{fig:ranking_reversals} shows the resulting ranking changes.

\begin{figure}[t]
    \centering
    \includegraphics[width=\linewidth]{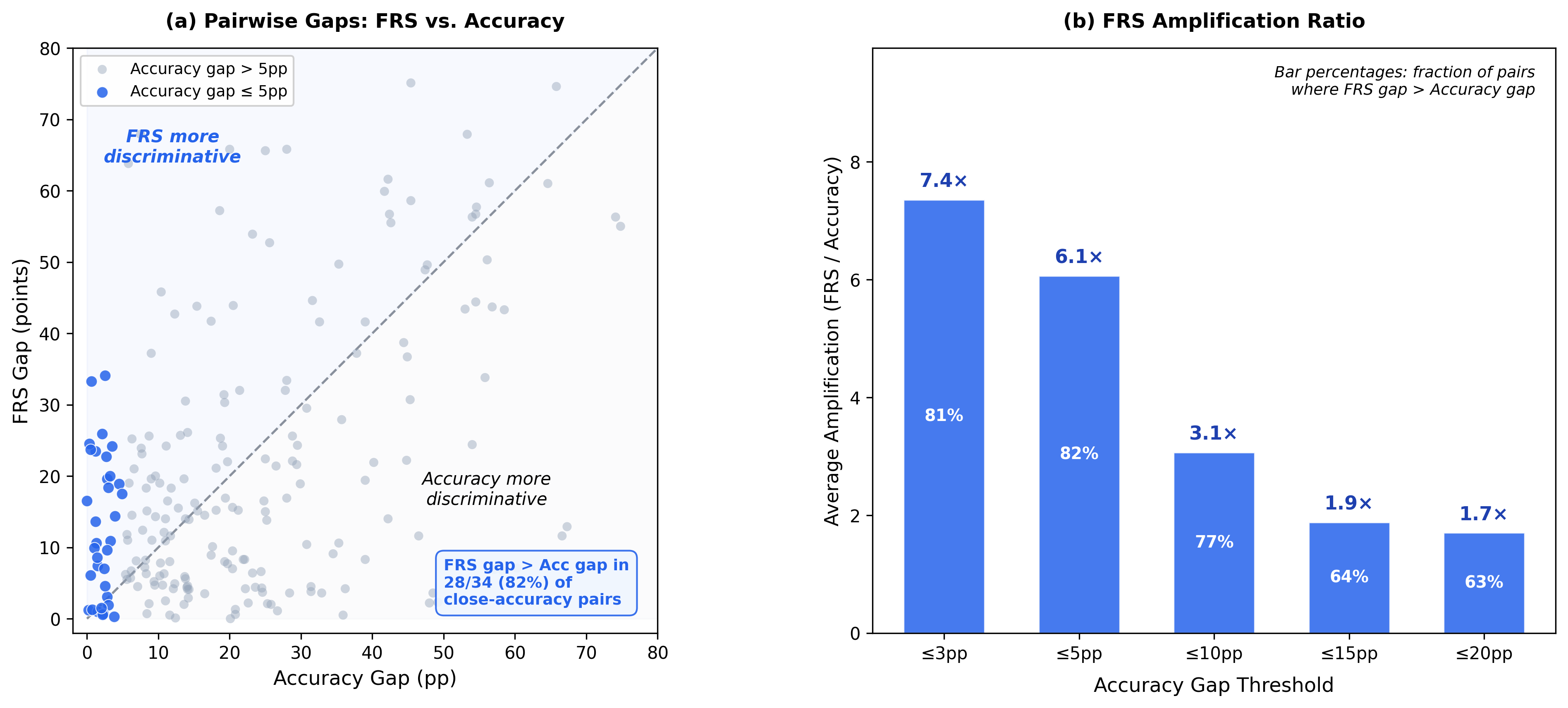}
    \caption{(a) Pairwise gaps: FRS vs.\ accuracy for all 216 per-benchmark model pairs. Blue points denote the 34 pairs with accuracy gap $\leq$5\,pp; FRS produces a larger gap in 82\% (28 of 34) of these close-accuracy cases. (b) FRS amplification ratio (mean FRS gap\,/\,mean accuracy gap) at increasing accuracy-gap thresholds. Bar percentages show the fraction of pairs where FRS gap exceeds accuracy gap. FRS is most discriminative where accuracy is least informative.}
    \label{fig:frs_amplification}
\end{figure}

\paragraph{FRS is most informative where accuracy is least discriminative.}
Among the 34 per-benchmark model pairs with greedy accuracy gaps of at most 5\,pp (out of 216 total), FRS produces a larger separation in 82\% of cases (28 of 34), with a mean gap roughly 6$\times$ greater than the corresponding accuracy gap. Among pairs within 3\,pp, the amplification rises to 7.4$\times$ (Figure~\ref{fig:frs_amplification}). Concrete examples confirm this: on MATH, DS-R1-7B and Qwen2.5-Math are exactly tied at 63.6\% greedy accuracy yet differ by 16.5 FRS points (Figure~\ref{fig:head_to_head}, Appendix~\ref{app:head_to_head}). Similarly, at the aggregate level DS-R1-7B and Phi-4-Reasoning are nearly tied on pass@1 yet differ by 18.8 FRS points.

\paragraph{Top-confidence accuracy is still not enough.}
A natural objection is to simply measure accuracy on the most-confident traces. This does not suffice. Phi-4-Reasoning achieves the highest top-10\% accuracy (79.3\%), slightly above DS-R1-7B (77.1\%), yet their FRS scores differ sharply: 69.7 vs.\ 88.5. The reason is that Phi-4-Reasoning's confident traces often contain correct answers wrapped in degenerate reasoning: short solution cores followed by repetitive continuations that inflate token-level confidence while degrading reasoning quality (Appendix~\ref{app:phi4_traces}).

\paragraph{FRS reshuffles accuracy-based rankings.} The per-benchmark effects above compound into substantial ranking shifts at the model level. Qwen2.5-7B, the highest-accuracy model overall at 73.5\%, drops from \#1 to \#7 under FRS, indicating that its confidence does not reliably prioritize its best reasoning. DS-R1-1.5B moves in the opposite direction, rising from \#8 by accuracy to \#2 under FRS despite having the second-lowest accuracy (42.9\%), because it consistently aligns high confidence with strong reasoning. All key ranking reversals involve FRS gaps exceeding 8 points with non-overlapping bootstrap 95\% confidence intervals.

These conclusions are robust across filtering thresholds, confidence estimators, judge models, and sampling budgets. Rankings are nearly unchanged under full-trace mean log-probability ($\rho \geq 0.983$) and remain strongly correlated under self-consistency, a logit-free estimator ($\rho = 0.80$; Appendix~\ref{app:confidence_proxies}). Full ablations are in Appendices~\ref{app:judge}, \ref{app:lowprob}, and \ref{app:k_sensitivity}.

\begin{table}[t]
\centering
\small
\caption{FRS at 10\% filtering (reasoning quality, 0--100) across 6 benchmarks. Acc Avg is greedy decoding pass@1 accuracy. Models are sorted by average FRS. $\pm$ values denote bootstrap standard deviation of the mean (10{,}000 resamples over 50 scored traces per bin); sensitivity to the upstream sampling budget is analyzed in Appendix~\ref{app:k_sensitivity}.}
\label{tab:frs_main}
\resizebox{\linewidth}{!}{%
\begin{tabular}{@{}l cccccc c c@{}}
\toprule
\textbf{Model} & \textbf{GSM8K} & \textbf{MATH} & \textbf{SVAMP} & \textbf{AQuA} & \textbf{GPQA} & \textbf{CSQA} & \textbf{FRS Avg} & \textbf{Acc Avg} \\
\midrule
DS-R1-7B     & 98.9{\scriptsize$\pm$1.1} & 94.0{\scriptsize$\pm$2.0} & 99.0{\scriptsize$\pm$0.7} & 96.8{\scriptsize$\pm$1.3} & 62.9{\scriptsize$\pm$2.5} & 79.2{\scriptsize$\pm$3.3} & \textbf{88.5}{\scriptsize$\pm$0.8} & 68.2 \\
DS-R1-1.5B   & 95.1{\scriptsize$\pm$2.2} & 84.5{\scriptsize$\pm$3.6} & 96.9{\scriptsize$\pm$1.7} & 90.1{\scriptsize$\pm$2.5} & 56.6{\scriptsize$\pm$4.1} & 56.1{\scriptsize$\pm$4.2} & \textbf{79.9}{\scriptsize$\pm$1.3} & 42.9 \\
Qwen3-4B     & 90.9{\scriptsize$\pm$2.3} & 63.5{\scriptsize$\pm$2.7} & 91.0{\scriptsize$\pm$1.9} & 78.5{\scriptsize$\pm$3.2} & 58.8{\scriptsize$\pm$2.2} & 75.0{\scriptsize$\pm$2.5} & \textbf{76.3}{\scriptsize$\pm$1.0} & 64.8 \\
Phi-4        & 91.5{\scriptsize$\pm$2.7} & 74.4{\scriptsize$\pm$4.3} & 75.5{\scriptsize$\pm$4.2} & 78.5{\scriptsize$\pm$3.5} & 48.4{\scriptsize$\pm$3.4} & 77.2{\scriptsize$\pm$3.8} & \textbf{74.2}{\scriptsize$\pm$1.5} & 62.5 \\
Qwen2.5-Math & 90.8{\scriptsize$\pm$2.5} & 77.5{\scriptsize$\pm$4.1} & 74.5{\scriptsize$\pm$4.4} & 74.9{\scriptsize$\pm$4.1} & 54.6{\scriptsize$\pm$3.0} & 70.6{\scriptsize$\pm$3.4} & \textbf{73.8}{\scriptsize$\pm$1.5} & 56.4 \\
Phi-4-Reas.  & 84.5{\scriptsize$\pm$2.6} & 80.0{\scriptsize$\pm$3.5} & 74.8{\scriptsize$\pm$3.4} & 83.2{\scriptsize$\pm$3.0} & 62.3{\scriptsize$\pm$4.9} & 33.4{\scriptsize$\pm$5.1} & \textbf{69.7}{\scriptsize$\pm$1.6} & 68.9 \\
Qwen2.5-7B   & 65.6{\scriptsize$\pm$2.2} & 75.6{\scriptsize$\pm$2.6} & 64.9{\scriptsize$\pm$1.9} & 77.2{\scriptsize$\pm$2.4} & 55.1{\scriptsize$\pm$2.7} & 70.1{\scriptsize$\pm$3.5} & \textbf{68.1}{\scriptsize$\pm$1.1} & 73.5 \\
LLaMA-3.1-8B & 79.9{\scriptsize$\pm$3.4} & 60.6{\scriptsize$\pm$4.3} & 80.0{\scriptsize$\pm$3.2} & 52.9{\scriptsize$\pm$3.9} & 50.5{\scriptsize$\pm$2.7} & 55.0{\scriptsize$\pm$3.0} & \textbf{63.1}{\scriptsize$\pm$1.4} & 59.3 \\
Gemma-7B     & 41.2{\scriptsize$\pm$5.1} & 18.9{\scriptsize$\pm$2.0} & 31.1{\scriptsize$\pm$4.6} & 22.2{\scriptsize$\pm$2.1} & 30.9{\scriptsize$\pm$1.7} & 13.4{\scriptsize$\pm$3.2} & \textbf{26.3}{\scriptsize$\pm$1.4} & 24.8 \\
\bottomrule
\end{tabular}
}
\end{table}

\paragraph{FRS predicts whether confidence-based selection helps or hurts.}
Beyond producing different rankings, FRS is the only metric that predicts a deployment-relevant outcome. For 54 model--benchmark pairs, we compare the reasoning quality of a model's top-confidence trace against a random baseline (Appendix~\ref{app:selection_gain}). FRS significantly predicts this \emph{selection gain} ($r{=}0.49$, $p{<}0.001$); pass@1, high-confidence accuracy, and unfiltered reasoning score are all nonsignificant ($p > 0.14$). The result is not an artifact of judge coupling: rescoring the selection-gain outcome with a held-out, different-family judge (Claude Haiku 4.5) preserves the effect ($r{=}0.367$, 95\% CI $[0.13, 0.56]$; Appendix~\ref{app:heldout_judge}), and it is not reducible to response style, since FRS remains predictive after jointly controlling for trace length, lexical diversity, and 4-gram repetition (partial $r{=}0.308$, $p{=}0.023$; Appendix~\ref{app:style_controls}).

\paragraph{Conditioning on confidence is what produces the signal.}
Confidence-agnostic evaluation of reasoning quality is not sufficient. We compare FRS against two confidence-agnostic baselines under the same judge: a single random trace per problem and the unfiltered average over all traces (Appendix~\ref{app:confidence_ablations}). Both amplify close-accuracy gaps at roughly 4$\times$ and neither predicts selection gain ($r{=}0.01$ and $r{=}0.008$, both n.s.), whereas FRS amplifies at 7.4$\times$ and predicts selection gain at $r{=}0.49$; bootstrap intervals over the $\leq$3\,pp amplification are disjoint ($[7.04, 8.82]$ vs.\ $[3.46, 5.64]$). Evaluating only the single most-confident trace, the $K{\to}1$ limit of confidence-conditioned evaluation, largely agrees with FRS (Spearman $\rho = 0.75$, amplification 7.9$\times$), with FRS providing a lower-variance estimate of the same quantity. This confirms that FRS captures a property that no confidence-agnostic metric measures: whether confidence-based selection actually surfaces better reasoning.

\begin{figure}[t]
\centering
\includegraphics[width=0.8\linewidth]{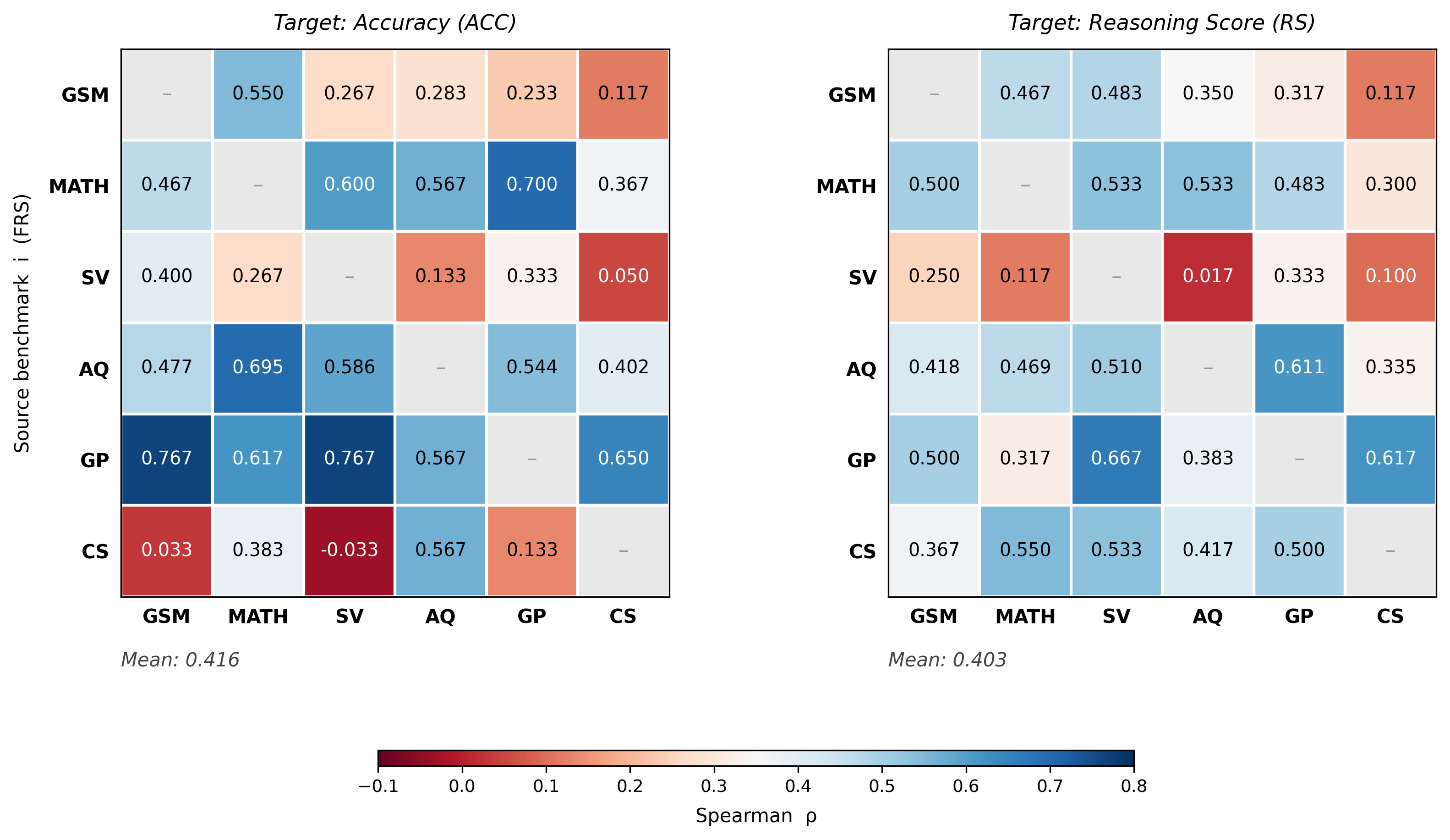}
\caption{Spearman correlations $\rho(\mathrm{FRS}_i, \cdot_j)$ across models for $i \neq j$. Rows are source benchmark $i$ (FRS), columns are target benchmark $j$. FRS has a mean $\rho = 0.416$ and $\rho = 0.403$ with Pass@1 (Accuracy) and Reasoning Score, respectively.}
\label{tab:frs_cross_benchmark}
\end{figure}

\paragraph{FRS correlates across benchmarks.}
Models with higher FRS on one benchmark tend to have higher accuracy and reasoning quality on other benchmarks. Figure~\ref{tab:frs_cross_benchmark} reports pairwise Spearman correlations $\rho(\mathrm{FRS}_i, \cdot_j)$ across models for $i \neq j$, with grand off-diagonal means of $\rho = 0.416$ (vs.\ accuracy) and $\rho = 0.403$ (vs.\ reasoning score). GPQA and AQuA yield the strongest row-averages (up to $\rho = 0.674$), while SVAMP and CSQA produce weaker correlations (as low as $\rho = 0.163$), suggesting that the strength of cross-benchmark transfer varies by domain.

In addition, we conduct a leave-one-benchmark-out (LOBO) analysis: for each held-out benchmark, we average each model's FRS over the remaining five benchmarks and correlate the result with the held-out FRS. Because DS-R1-7B and DS-R1-1.5B share a distillation training process and may produce correlated FRS across benchmarks, we report a family-controlled variant, holding out both DS-R1 models jointly, as the primary number: the mean Spearman correlation is $\rho = 0.495$ across the remaining 7 models, positive in all six folds. Without the family control, the mean correlation over all 9 models is $\rho = 0.712$ (std.\ 0.158), with all six folds positive and five of six significant by permutation test ($p < 0.05$). The strongest generalization appears for GSM8K ($\rho = 0.933$, $p < 0.001$) and AQuA ($\rho = 0.820$, $p = 0.004$); the weakest is CSQA ($\rho = 0.467$, $p = 0.102$), consistent with commonsense reasoning placing different demands on confidence--quality alignment. These results indicate partial, not perfect, transfer: confidence--quality alignment carries cross-benchmark structure that is invisible to accuracy, and knowing a model's FRS on five benchmarks is informative about its FRS on the sixth.

\section{Discussion}
\label{sec:discussion}

\paragraph{FRS as a practical audit for confidence-based selection.}
Many deployment strategies (e.g., abstention and test-time compute scaling) implicitly assume that a model's highest-confidence outputs are also its best ones. Our results show this does not hold uniformly. FRS provides a simple pre-deployment audit: if FRS improves under tighter filtering, confidence-based selection will likely surface strong reasoning; if it degrades, the same strategy may surface weaker reasoning at precisely the moments the model appears most certain. The cost is modest: rankings at $k{=}8$ traces per problem remain nearly identical to those at $k{=}16$ (Appendix~\ref{app:k_sensitivity}). As shown in Section~\ref{sec:frs_results}, this holds across confidence estimators, suggesting confidence--quality alignment is a model property rather than an artifact of aggregation.

\paragraph{FRS fills a gap in the evaluation stack.}
Accuracy measures what a model gets right, but not whether the reasoning behind it is sound. Correct answers can arise from flawed or degenerate traces, so outcome-based metrics alone cannot assess reasoning quality in the high-confidence region where deployed systems operate. As our selection gain analysis confirms (Section~\ref{sec:frs_results}), this gap is deployment-relevant: FRS is the only metric among six candidates that significantly predicts whether confidence-based selection improves or degrades reasoning quality, a result that replicates under a held-out different-family judge (Appendix~\ref{app:heldout_judge}) and diverges from what process reward models capture over the full trace distribution (Appendix~\ref{app:prm}). Moreover, our leave-one-benchmark-out analysis shows that confidence--quality alignment carries stable cross-benchmark structure (family-controlled $\rho = 0.495$ across held-out folds, positive in all folds; $\rho = 0.712$ without the family control), not a benchmark-specific artifact. Together, these results establish confidence-conditioned reasoning quality as a distinct evaluation target for reasoning models in confidence-mediated settings.

\paragraph{Scope and limitations.}
FRS targets the broad class of confidence-mediated selection: best-of-N by self-certainty, confidence-based abstention, and confidence-thresholded test-time scaling. It does not assume that high-confidence traces are always the deployment-relevant outputs; rather, that premise is exactly what FRS audits, and Appendix~\ref{app:conf_distributions} shows models where the confidence--correctness relationship is inverted. Inference pipelines built on verifier reranking, debate, or search change only the selection rule, not the rubric: FRS can be applied to whichever trace such a pipeline selects, and self-consistency, itself a confidence signal, is already covered (Appendix~\ref{app:self_consistency}). Our evaluation spans arithmetic, competition math, algebraic word problems, graduate-level science, and commonsense reasoning; demonstrating confidence--quality alignment directly on agentic planning, interactive tool use, code reasoning, and long-horizon decision-making remains an important direction, and the long-horizon case is the regime our motivation most directly targets, where low-confidence correct traces are most likely coincidental. Finally, the low-probability confidence estimator can be inflated by degenerate repetition (Appendix~\ref{app:phi4_traces}); we treat this as a limitation of the confidence signal that the rubric-based judge corrects, and note that FRS is an evaluation metric rather than a reward signal, so exploiting this artifact would require optimization pressure the evaluation setting lacks.

\paragraph{What drives confidence--quality alignment?}
One striking pattern is that models whose FRS improves under tighter filtering all used reinforcement learning during post-training, whereas models whose FRS degrades did not. We view this as suggestive rather than conclusive: it is consistent with prior findings that RL improves confidence calibration, but our experiments were not designed to isolate RL from architecture, scale, or training data. We treat confidence--quality alignment as an important phenomenon revealed by FRS, and its mechanisms as an open question. If this alignment can be trained directly, FRS may serve not only as an evaluation metric but as a training target for improving reasoning models.


\section{Conclusion}
\label{sec:conclusion}

We introduced the Filtered Reasoning Score (FRS), a metric that evaluates reasoning quality on a model's most-confident outputs. Conditioning evaluation on confidence reveals structure that accuracy cannot: models indistinguishable under accuracy, and even under top-confidence accuracy, can differ sharply in the reasoning their confidence selects. FRS is most discriminative exactly where accuracy is least informative.

Evaluation should move beyond correctness when models are deployed in confidence-mediated settings. Accuracy measures whether a model can produce the right answer, but not whether its most-confident outputs reflect sound reasoning. FRS targets this question directly, and our leave-one-benchmark-out analysis confirms that the reasoning property it measures, confidence--quality alignment, partially transfers across benchmarks even under family-controlled evaluation.

An important direction for future work is understanding and improving confidence--quality alignment. Our results show this alignment varies systematically across models and training regimes, raising the possibility of treating confidence-conditioned reasoning quality not only as an evaluation target but as a training objective, encouraging models to assign high confidence selectively to well-reasoned outputs.
\newpage

\section*{Ethics Statement}

This work introduces the \textbf{Filtered Reasoning Score (FRS)}, an evaluation metric designed to assess \emph{confidence-conditioned reasoning quality}. We view FRS as an auditing tool for modern deployment settings where systems act on a model’s most-confident outputs. In such settings, correctness alone is insufficient: models can produce correct answers through flawed or degenerate reasoning, and confidence can preferentially surface these failures. By explicitly measuring reasoning quality under confidence-based selection, FRS aims to reduce misplaced trust in high-confidence outputs and improve the reliability of model evaluation.

The primary positive impact of FRS is to expose failure modes that are invisible to accuracy and even to top-confidence accuracy. By identifying whether a model’s confidence aligns with strong reasoning, FRS enables practitioners to make more informed decisions about model selection, filtering strategies, and deployment policies. We therefore position FRS as a practical pre-deployment diagnostic: when FRS is high, confidence-based selection is more likely to surface sound reasoning; when FRS is low, such strategies may amplify poor reasoning despite high apparent confidence.

At the same time, FRS is not a complete measure of reliability or safety. A model with high FRS may still be unsafe, biased, or poorly calibrated in real-world settings. In addition, because FRS relies on LLM-based judges, it may inherit biases from the judge model and favor reasoning styles that are more legible to the evaluator. Like any metric, FRS may also be subject to optimization or gaming. We therefore emphasize that FRS should be used as \emph{one component} of a broader evaluation stack, alongside accuracy, calibration, robustness testing, and human oversight, rather than as a standalone deployment criterion.

We take steps to mitigate these risks. We validate FRS across judge models and human annotations, report robustness across filtering thresholds and sampling budgets, and open-source the FRS evaluation pipeline to promote transparency and reproducibility. We also encourage users to evaluate FRS across diverse benchmarks and domains, as confidence--quality alignment may vary across settings.

FRS also raises fairness considerations. Benchmarks, prompts, and judge models may encode narrow distributions of tasks or linguistic norms, which can affect FRS scores across models or domains. As a result, FRS-based rankings should not be used in isolation for socially consequential decisions (e.g., in healthcare, education, or legal contexts). Instead, FRS should serve as a diagnostic signal within a broader, human-centered evaluation process.

Our experiments primarily use public benchmarks and model-generated outputs, and do not require collection of personal user data. For the human annotation study used to validate FRS (Appendix~\ref{app:human_annotation}), all 10 annotators were informed of the study's purpose and how their annotations would be used, and provided consent before participating; annotation was conducted blind to the automated judge's scores, no personal or sensitive data was collected from annotators, and annotators participated voluntarily. We also respect dataset and model licenses and aim to ensure reproducibility through release of the FRS codebase.

Finally, we note that FRS requires multi-sample generation and evaluation, which incurs additional computational cost compared to accuracy. In practice this cost is modest: the full main evaluation across all 9 models and 6 benchmarks requires 13{,}500 judged traces at a total judging cost of approximately \$12 with GPT-4o-mini, and FRS requires no generation beyond standard best-of-N pipelines, since it operates on traces already produced during inference (Appendix~\ref{app:cost}). We therefore report efficiency trade-offs and show that FRS rankings remain stable under reduced sampling. We view this cost as justified by the improved visibility FRS provides into how confidence interacts with reasoning quality. Overall, FRS contributes toward more responsible evaluation by shifting attention from whether models are correct to whether they are reliable \emph{when they appear most confident}.

\bibliography{colm2026_conference}
\bibliographystyle{plainnat}


\newpage
\appendix

\section{Scoring Rubric for Reasoning Quality}
\label{app:rubric}

We provide the full prompt given to the GPT-4o-mini judge for scoring reasoning traces.

\begin{promptbox}[REASONING QUALITY SCORING PROMPT]
You are an expert evaluator of mathematical and logical reasoning. Score the chain-of-thought (CoT) on 4 dimensions using the scoring criteria below.

Each score must be an integer from 1--5 (1 = very poor, 5 = excellent).

\tcblower

\textbf{1.\ FAITHFULNESS (1--5)} \\
\textit{Definition:} Reasoning is internally consistent, follows logical rules, and stays focused on the problem without hidden shortcuts or leaps.

\begin{itemize}[nosep, leftmargin=1em]
\item \textbf{5:} Perfect logical consistency, no contradictions, stays completely on-topic
\item \textbf{4:} Minor inconsistencies or slight tangents, but overall coherent
\item \textbf{3:} Some logical gaps or moderate off-topic content
\item \textbf{2:} Significant logical flaws or frequent tangents
\item \textbf{1:} Major contradictions, illogical leaps, or completely off-topic
\end{itemize}

\textit{Dock points for:} Contradictory statements, logical leaps without justification, going off-topic, hidden assumptions, unjustified final answers, shortcut reasoning.

\vspace{6pt}
\textbf{2.\ UTILITY (1--5)} \\
\textit{Definition:} Each step meaningfully contributes to solving the problem, calculations are correct, and reasoning efficiently leads to the final answer.

\begin{itemize}[nosep, leftmargin=1em]
\item \textbf{5:} Every step is necessary and correct, efficient path to solution
\item \textbf{4:} Most steps useful, minor inefficiencies or small errors
\item \textbf{3:} Some useful steps mixed with unnecessary ones or calculation errors
\item \textbf{2:} Many unnecessary steps or significant calculation errors
\item \textbf{1:} Mostly useless steps, major calculation errors, or repetitive content
\end{itemize}

\textit{Dock points for:} Incorrect calculations, repetitive statements, unnecessary verbose explanations, steps that don't advance toward the solution, circular logic.

\vspace{6pt}
\textbf{3.\ COHERENCE (1--5)} \\
\textit{Definition:} Steps flow smoothly from one to the next with clear logical progression and smooth transitions.

\begin{itemize}[nosep, leftmargin=1em]
\item \textbf{5:} Perfect flow, each step naturally follows from the previous
\item \textbf{4:} Good flow with minor awkward transitions
\item \textbf{3:} Some disjointed steps but overall progression
\item \textbf{2:} Choppy flow with unclear connections between steps
\item \textbf{1:} Disjointed, random steps with no clear progression
\end{itemize}

\textit{Dock points for:} Abrupt transitions, missing connecting logic, disjointed sequence, poor organization, dangling references, disordered reasoning chain.

\vspace{6pt}
\textbf{4.\ FACTUALITY (1--5)} \\
\textit{Definition:} Every step must be factually correct and grounded in the problem context, not hallucinated from surface-level understanding.

\begin{itemize}[nosep, leftmargin=1em]
\item \textbf{5:} All facts and statements are accurate and grounded in the problem
\item \textbf{4:} Mostly accurate with minor factual errors
\item \textbf{3:} Some factual errors or unsupported claims
\item \textbf{2:} Multiple factual errors or significant hallucinations
\item \textbf{1:} Major factual errors, hallucinations, or completely unsupported claims
\end{itemize}

\textit{Dock points for:} Hallucinated facts, incorrect interpretations, unsupported assumptions, surface-level understanding leading to wrong facts, claims contradicting problem evidence.

\vspace{6pt}
\textbf{EVALUATION PROCESS}
\begin{enumerate}[nosep, leftmargin=1.5em]
\item Read the problem carefully to understand the context and given information
\item Analyze each step of the CoT reasoning
\item Check each step against the four criteria above
\item Assign scores based on the specific guidelines for each dimension
\item Ensure every step is evaluated for factual accuracy and logical soundness
\end{enumerate}

\vspace{4pt}
\textbf{Input fields:} \texttt{\{problem\}}, \texttt{\{cot\}}, \texttt{\{gold\}}, \texttt{\{flags\_summary\}}, \texttt{\{evidence\}}

\vspace{4pt}
\textbf{Required output (JSON):}\\
\texttt{\{"faithfulness": <1-5>, "utility": <1-5>, "coherence": <1-5>, "factuality": <1-5>\}}
\end{promptbox}

\FloatBarrier
\section{Rubric Dimension Analysis}
\label{app:dim_ablation}

A natural question is whether the four rubric dimensions capture distinct information or are functionally redundant. We analyze the dimension structure across all 54 model--benchmark pairs.

\paragraph{Pairwise structure.} The four dimensions organize into two axes. Faithfulness, coherence, and utility are strongly mutually correlated (Spearman $\rho \in [0.90, 0.98]$), forming a dominant logical-quality axis. Factuality is largely independent of this axis, correlating at $\rho \in [0.35, 0.61]$ with each of the other three dimensions, and contributes a distinct grounding signal. Consistent with this structure, the model-level FRS ranking induced by faithfulness alone correlates with the full four-dimension ranking at $\rho = 0.994$, with coherence and utility behaving similarly ($\rho \geq 0.99$), while factuality alone yields $\rho = 0.861$ and produces meaningful ranking shifts when included.

\paragraph{Why factuality differs from faithfulness in practice.} The two dimensions penalize different failure modes by construction (Appendix~\ref{app:rubric}). Faithfulness penalizes internal inconsistency: contradictions, unjustified leaps, and hidden shortcuts, regardless of whether the trace's premises are grounded in the problem. Factuality penalizes hallucinated premises, misread problem context, and unsupported claims, even when the subsequent chain is internally consistent. On the largely arithmetic benchmarks in our suite these failure modes frequently co-occur, since an error that breaks grounding usually also breaks internal consistency, which is why the logical-quality axis dominates. The dimensions nevertheless dissociate in both directions: on knowledge-heavy benchmarks such as GPQA, a trace can be internally coherent while resting on a hallucinated fact, and the Phi-4-Reasoning traces in Appendix~\ref{app:phi4_traces} exhibit the converse pattern directly, scoring 5 on factuality while faithfulness and utility collapse to 2 or below, because every claim is grounded in the problem even though the reasoning degenerates into repetition.

\paragraph{Leave-one-dimension-out ablation.} We recompute the paper's two headline quantities, close-accuracy amplification (Section~\ref{sec:frs_results}) and LOBO transfer, under each three-dimension rubric subset.

\begin{table}[!ht]
\centering\small
\caption{FRS behavior under leave-one-dimension-out rubric subsets. Amplification is the mean FRS gap over the mean accuracy gap among model pairs within the stated accuracy threshold; Transfer is the mean LOBO Spearman $\rho$.}
\label{tab:dim_ablation}
\begin{tabular}{@{}l ccc@{}}
\toprule
\textbf{Rubric Subset} & \textbf{Amp.\ $\leq$3\,pp} & \textbf{Amp.\ $\leq$5\,pp} & \textbf{Transfer LOBO $\rho$} \\
\midrule
Full (4 dimensions) & 7.4$\times$ & 6.1$\times$ & 0.71 \\
Drop faithfulness   & 7.5$\times$ & 6.2$\times$ & 0.71 \\
Drop coherence      & 6.6$\times$ & 5.5$\times$ & 0.62 \\
Drop utility        & 6.5$\times$ & 5.4$\times$ & 0.63 \\
Drop factuality     & 9.5$\times$ & 7.9$\times$ & 0.75 \\
\bottomrule
\end{tabular}
\end{table}

FRS is stable to removing any single logical-axis dimension, with amplification in the 6.5--7.5$\times$ range and transfer in the 0.62--0.71 range, so no individual dimension is load-bearing. Dropping factuality \emph{raises} amplification to 9.5$\times$ because grounding errors are rare on these largely arithmetic benchmarks, so removing the factuality signal sharpens separation along the logical-quality axis. We nevertheless retain all four dimensions following the taxonomy of \citet{lee2025evaluating}: factuality is the one axis carrying independent information ($\rho = 0.861$ vs.\ $\geq 0.99$ for the others), and it is the relevant signal for knowledge-heavy benchmarks and deployment auditing, where hallucinated-but-coherent reasoning is precisely the failure mode outcome metrics miss.

\FloatBarrier
\section{Low-Probability Token Analysis}
\label{app:lowprob}

\begin{figure}[!ht]
    \centering
    \includegraphics[width=0.9\linewidth]{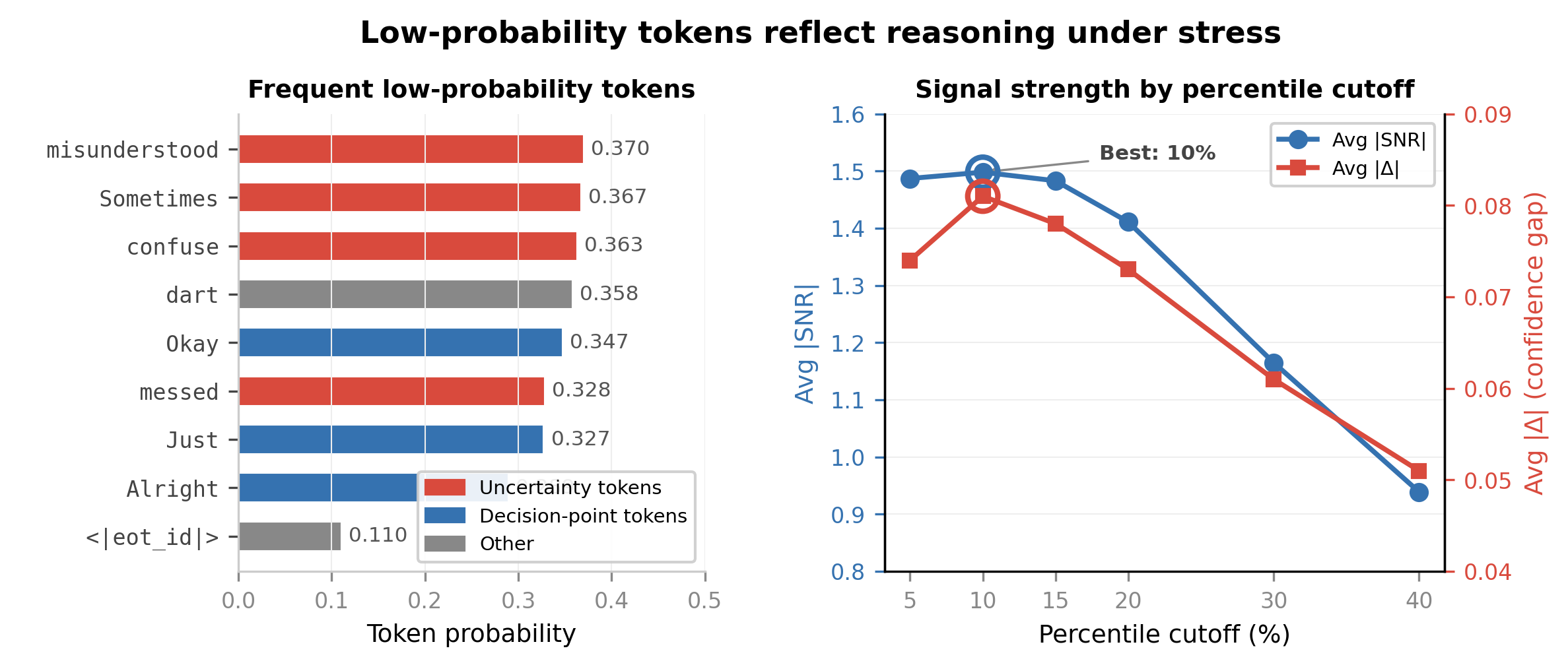}
    \caption{Left: Frequent low-probability tokens, categorized as uncertainty-expressing (red) or decision-point (blue). Right: Confidence signal quality (SNR and mean confidence gap) as a function of percentile cutoff. The 10\% cutoff yields the strongest signal on both measures.}
    \label{fig:low_prob_analysis}
\end{figure}

\paragraph{What are low-probability tokens?} The most frequent low-probability tokens (Figure~\ref{fig:low_prob_analysis}, left) fall into two categories:

\begin{enumerate}[nosep]
    \item \textbf{Decision-point tokens}: Words that are largely interchangeable without affecting reasoning logic, such as ``Okay,'' ``Alright,'' and ``Just.'' Low probability at these tokens reflects a choice among multiple equally valid continuations, indicating a natural branch point in the generation process. Low probability at these tokens is frequent and normal across models.
    \item \textbf{Uncertainty-expressing tokens}: Words that explicitly signal confusion or doubt, such as ``messed,'' ``confuse,'' ``misunderstood,'' and ``Sometimes.'' Low probability at these tokens reflects genuine uncertainty in the reasoning process.
\end{enumerate}

\noindent Both categories represent points where the model's reasoning is under stress, either because multiple paths are available or because the model is uncertain about how to proceed.

\paragraph{Analyzing the percentile cutoff.} We evaluate percentile cutoffs from 5\% to 40\% using two separation metrics across all 9 models and 6 benchmarks. The first is the signal-to-noise ratio (SNR), defined as:
\begin{equation}
    \text{SNR} = \frac{\mathbb{E}[C \mid Y{=}1] - \mathbb{E}[C \mid Y{=}0]}{\sqrt{\frac{1}{2}\left(\text{Var}[C \mid Y{=}1] + \text{Var}[C \mid Y{=}0]\right)}}
\end{equation}
with $Y_i \in \{0,1\}$ denoting correctness. We also compute the average absolute difference in mean confidence between correct and incorrect populations (Figure~\ref{fig:low_prob_analysis}, right).

The lowest 5--15\% of tokens yield the strongest separation, with the 10\% cutoff achieving the highest values on both measures (Avg $|$SNR$|$ = 1.498, Avg $|\Delta|$ = 0.081). Beyond 15\%, both metrics decline monotonically as high-probability tokens dilute the signal. The 5--15\% range is consistently strong; we use 10\% as the default throughout the paper because it provides the best overall signal while including enough tokens per trace for a stable estimate.

\section{LLM Judge Validation}
\label{app:judge}

We validate the GPT-4o-mini judge through two complementary approaches: cross-model agreement with independent LLM validators, and a human annotation study.

\subsection{Cross-Model Agreement}

We evaluate 500 stratified samples spanning all 9 models, 6 datasets, and 4 reasoning dimensions using two independent validators: GPT-4o and Claude Sonnet 4.5 (Figure~\ref{fig:judge_agreement}).

GPT-4o-mini agrees with GPT-4o within $\pm$1 point for 82\% of scores (50\% exact match). Cross-family agreement with Claude Sonnet 4.5: 75\% within $\pm$1 (41\% exact match).

\begin{center}
    \includegraphics[width=\linewidth]{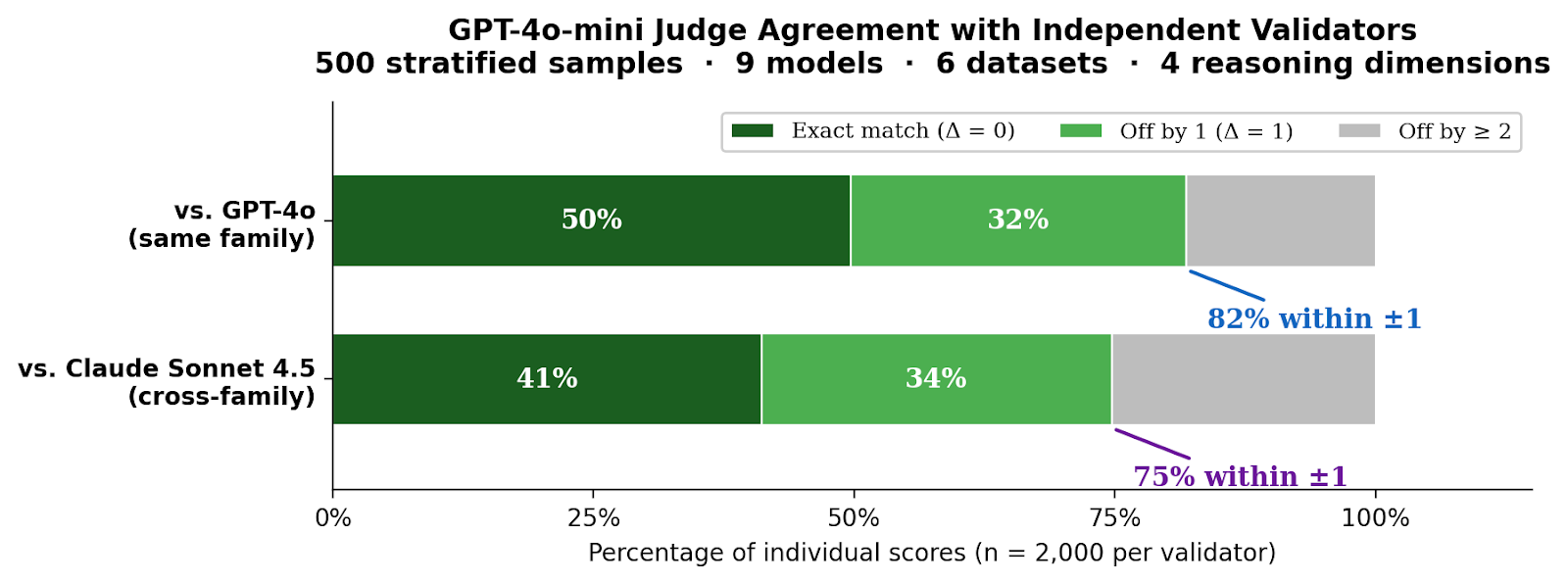}
    \captionof{figure}{GPT-4o-mini judge agreement with independent validators on 500 stratified samples (n = 2{,}000 per validator). Scores agree within $\pm$1 point for 82\% (vs.\ GPT-4o) and 75\% (vs.\ Claude Sonnet 4.5).}
    \label{fig:judge_agreement}
\end{center}

\subsection{Human Annotation Study}
\label{app:human_annotation}

To validate the GPT-4o-mini judge against human judgment, we conducted a human annotation study with 10 annotators. We collected 500 total annotations: each annotator independently labeled 50 samples, with no overlap between annotators (i.e., each sample was labeled by exactly one annotator).

\paragraph{Annotation protocol.} The study was conducted blind: annotators did not see GPT-4o-mini's scores. Each sample presented to an annotator consisted of: (1) the original question, (2) the model's full reasoning trace and output, and (3) the ground-truth answer. Annotators evaluated each response across the same four reasoning dimensions used throughout this work (faithfulness, utility, coherence, and factuality), each scored on the 1--5 integer scale defined in the rubric (Appendix~\ref{app:rubric}). The 500 samples were drawn via stratified sampling across all 9 models and 6 benchmarks to ensure broad coverage.

\paragraph{Agreement metrics.} We compute agreement between the GPT-4o-mini judge and human annotators at the level of individual dimension scores. Each of the 500 samples yields 4 dimension-level comparisons, producing 2{,}000 human--judge score pairs in total. Following the same convention used for cross-model validation, we define ``within $\pm$1'' as an absolute difference $\leq 1$ on the 1--5 scale.

\paragraph{Results.} GPT-4o-mini agrees with human annotators within $\pm$1 for 78\% of dimension-level scores, with a Spearman rank correlation of $\rho = 0.73$ across all 2{,}000 comparisons. The human--judge agreement (78\%) is slightly below the GPT-4o agreement (82\%) but above the cross-family Claude Sonnet 4.5 agreement (75\%), and the correlation of $\rho = 0.73$ indicates that the automated judge preserves the rank ordering of reasoning quality that human annotators assign.

\paragraph{Discussion.} The strong correlation ($\rho = 0.73$) confirms that GPT-4o-mini's scores track the relative quality judgments of human evaluators, supporting its use as the scoring function inside FRS. Taken together, the cross-model and human validation provide converging evidence that the automated judge is a reliable proxy for reasoning quality assessment.

\section{Convergence Analysis}
\label{app:convergence}

Figure~\ref{fig:convergence} (main text) shows the aggregate convergence trend. The per-benchmark breakdowns below confirm the pattern holds individually: reasoning scores exhibit lower variance at every sample size across all six benchmarks.

\begin{center}
    \includegraphics[width=\linewidth]{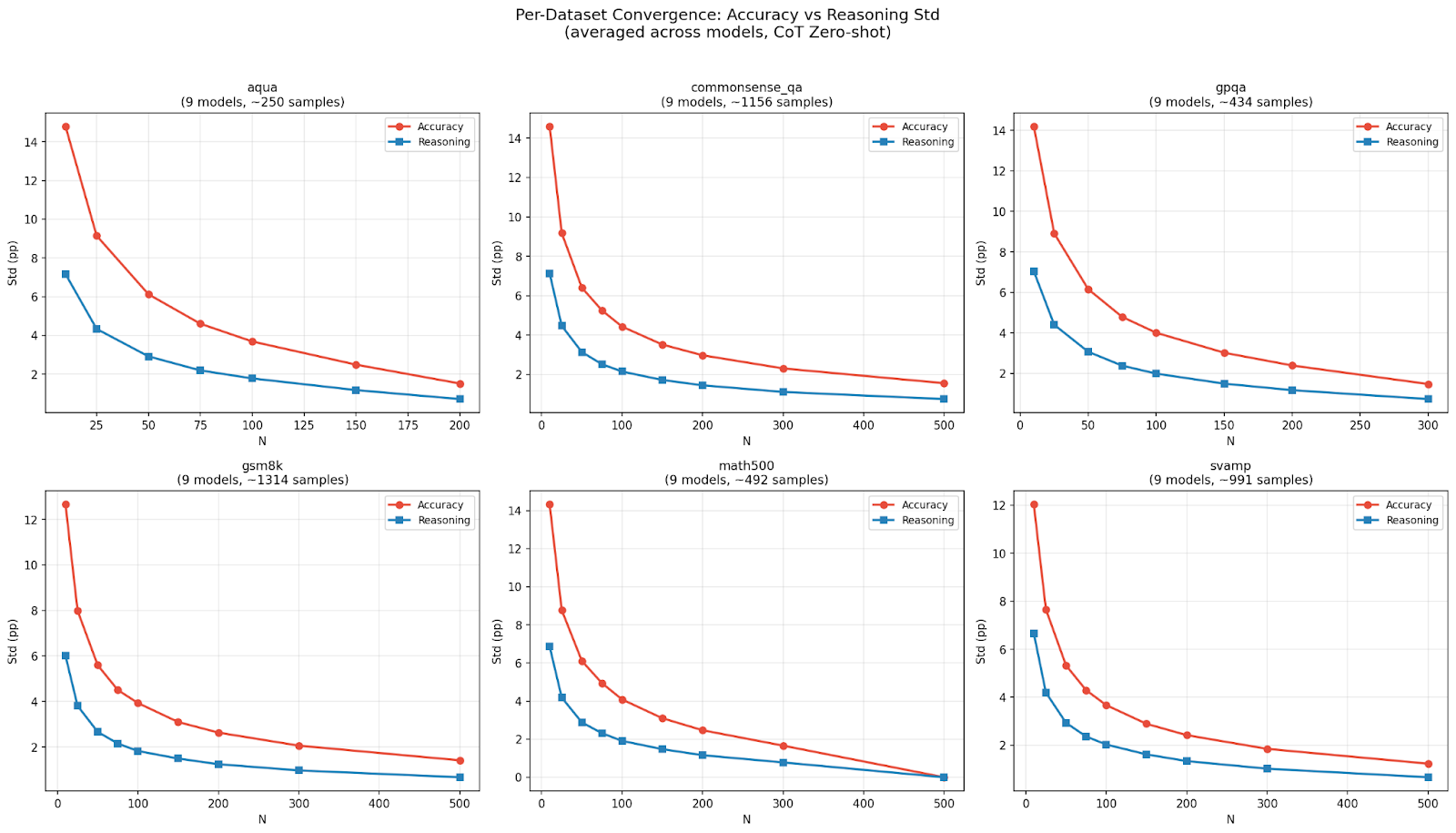}
    \captionof{figure}{Per-benchmark convergence of accuracy vs.\ reasoning score as a function of evaluation set size $N$, averaged across 9 models (CoT zero-shot). Reasoning scores converge faster and exhibit lower variance at all sample sizes on every benchmark.}
    \label{fig:convergence_perbench}
\end{center}

\section{Ranking Stability Across Evaluation Conditions}
\label{app:ranking_stability}

\begin{table}[!ht]
\centering
\small
\caption{Ranking stability across four evaluation conditions (direct, few-shot, CoT $T{=}0$, CoT $T{=}0.7$). Reasoning score produces more stable rankings on 5 of 6 benchmarks.}
\label{tab:ranking_stability}
\begin{tabular}{@{}l cc cc cc@{}}
\toprule
& \multicolumn{2}{c}{\textbf{Avg Spearman $\rho$ ($\uparrow$)}} & \multicolumn{2}{c}{\textbf{Rank inversions ($\downarrow$)}} & \multicolumn{2}{c}{\textbf{Avg max swing ($\downarrow$)}} \\
\cmidrule(lr){2-3} \cmidrule(lr){4-5} \cmidrule(lr){6-7}
\textbf{Benchmark} & \textbf{Reas.} & \textbf{Acc.} & \textbf{Reas.} & \textbf{Acc.} & \textbf{Reas.} & \textbf{Acc.} \\
\midrule
GSM8K       & \textbf{0.81} & 0.37 & \textbf{14.4\%} & 33.3\% & \textbf{1.9} & 3.9 \\
MATH500     & \textbf{0.85} & 0.63 & \textbf{12.5\%} & 25.0\% & \textbf{1.6} & 2.9 \\
SVAMP       & \textbf{0.76} & 0.26 & \textbf{19.4\%} & 38.4\% & \textbf{2.4} & 4.3 \\
AQuA        & 0.75 & \textbf{0.77} & 19.4\% & \textbf{19.0\%} & 2.3 & \textbf{2.3} \\
GPQA        & \textbf{0.36} & 0.24 & \textbf{36.1\%} & 39.4\% & \textbf{3.9} & 4.7 \\
CSQA        & \textbf{0.78} & 0.63 & \textbf{17.6\%} & 24.1\% & \textbf{2.3} & 3.0 \\
\midrule
\textbf{Average} & \textbf{0.72} & 0.49 & \textbf{19.9\%} & 29.9\% & \textbf{2.4} & 3.5 \\
\bottomrule
\end{tabular}
\end{table}

\section{Full Results Across Thresholds}
\label{app:fullresults}

This appendix provides the complete FRS and accuracy results across all filtering thresholds $K \in \{10, 20, 30, 40, 50\}$. DS-R1-7B ranks \#1 by average FRS at every threshold. Qwen2.5-7B ranks \#7 at every threshold. Spearman $\rho$ between FRS rankings at $K{=}10$ and other thresholds: 0.91 ($K{=}20$), 0.73 ($K{=}30$), 0.72 ($K{=}40$), 0.68 ($K{=}50$).

\begin{table}[!ht]
\centering\small
\caption{Average FRS (0--100) across 6 benchmarks at each filtering threshold $K$. Models sorted by FRS at $K{=}10\%$. Per-benchmark breakdowns available upon request.}
\label{tab:frs_all_thresholds}
\begin{tabular}{@{}l ccccc@{}}
\toprule
\textbf{Model} & $K{=}10\%$ & $K{=}20\%$ & $K{=}30\%$ & $K{=}40\%$ & $K{=}50\%$ \\
\midrule
DS-R1-7B       & \textbf{88.5} & 88.2 & 87.5 & 86.7 & 85.7 \\
DS-R1-1.5B     & \textbf{79.9} & 78.1 & 76.3 & 74.9 & 72.2 \\
Qwen3-4B       & \textbf{76.3} & 76.2 & 75.1 & 75.0 & 74.0 \\
Phi-4          & \textbf{74.2} & 72.8 & 72.4 & 72.8 & 73.4 \\
Qwen2.5-Math   & \textbf{73.8} & 75.8 & 76.8 & 77.1 & 76.9 \\
Phi-4-Reas.    & \textbf{69.7} & 76.2 & 79.7 & 81.1 & 81.5 \\
Qwen2.5-7B     & \textbf{68.1} & 68.6 & 67.9 & 67.0 & 67.5 \\
LLaMA-3.1-8B   & \textbf{63.1} & 60.1 & 58.4 & 58.5 & 58.7 \\
Gemma-7B       & \textbf{26.3} & 33.6 & 37.7 & 39.6 & 40.1 \\
\bottomrule
\end{tabular}
\end{table}

\FloatBarrier
\section{Composition of the Top-10\% Filtered Set}
\label{app:selected_composition}

Because FRS pools all $k{=}16$ traces across all problems within a model--benchmark pair and ranks them globally by confidence, a potential concern is that the top-10\% set disproportionately draws from problems where the model is uniformly confident, i.e., easy problems, so that FRS would measure reasoning quality on easy problems rather than on typical outputs. We test this directly by comparing the difficulty distribution of the selected traces against the benchmark distribution.

\paragraph{Problem coverage.} The filtered set is broad rather than concentrated: 63.8\% of problems contribute at least one trace to the top-10\% set (per-pair mean; 67\% macro-weighted across pairs), consistent with the coverage analysis in Appendix~\ref{app:k_sensitivity}.

\paragraph{Difficulty distribution.} Defining per-problem difficulty by the solve rate among the $k{=}16$ samples, the filtered set draws \emph{more} from hard problems than from uniformly easy ones: 22.3\% of selected traces come from problems solved at most 4 of 16 times, versus only 16.4\% from problems solved 16 of 16 times (per-pair means).

\paragraph{MATH500 difficulty levels.} MATH500 provides annotated difficulty levels 1--5, allowing a direct comparison of the selected-trace level distribution against the original benchmark distribution.

\begin{table}[!ht]
\centering\small
\caption{Share of traces (\%) by MATH500 difficulty level: top-10\% selected set vs.\ the original benchmark distribution.}
\label{tab:math500_levels}
\begin{tabular}{@{}l ccccc@{}}
\toprule
 & \textbf{Level 1} & \textbf{Level 2} & \textbf{Level 3} & \textbf{Level 4} & \textbf{Level 5} \\
\midrule
Selected (top 10\%) & 10.3 & 18.6 & 22.0 & 27.8 & 21.2 \\
Original            &  8.6 & 18.0 & 21.0 & 25.6 & 26.8 \\
\bottomrule
\end{tabular}
\end{table}

Selection mildly underweights only the hardest level rather than concentrating on easy problems: levels 4--5 still supply 49\% of selected traces, the two easiest levels gain only 2.3\,pp combined, and the only notable shift is a 5.6\,pp drop at level 5. We conclude that FRS at $K{=}10\%$ evaluates reasoning quality on a difficulty distribution close to the benchmark's own, not on an easy-problem subset.

\FloatBarrier
\section{Accuracy Across Confidence Thresholds}
\label{app:acc_thresholds}

Average accuracy on the top $K\%$ most confident traces, computed from the same pass@16 sampling runs used for FRS.

\begin{table}[!ht]
\centering\small
\caption{Average accuracy (\%) on the top $K\%$ most confident traces across 6 benchmarks. Models sorted by top-10\% accuracy. Per-benchmark breakdowns available upon request.}
\label{tab:acc_all_thresholds}
\begin{tabular}{@{}l ccccc@{}}
\toprule
\textbf{Model} & $K{=}10\%$ & $K{=}20\%$ & $K{=}30\%$ & $K{=}40\%$ & $K{=}50\%$ \\
\midrule
Phi-4-Reas.    & \textbf{79.3} & 81.5 & 81.8 & 81.5 & 80.8 \\
DS-R1-7B       & \textbf{77.1} & 76.3 & 75.8 & 75.4 & 74.6 \\
DS-R1-1.5B     & \textbf{70.2} & 67.8 & 65.5 & 63.8 & 61.8 \\
Qwen3-4B       & \textbf{68.0} & 66.7 & 65.8 & 65.0 & 64.2 \\
LLaMA-3.1-8B   & \textbf{64.6} & 60.1 & 58.2 & 57.5 & 57.0 \\
Qwen2.5-7B     & \textbf{64.5} & 63.1 & 62.2 & 60.9 & 60.8 \\
Phi-4          & \textbf{51.2} & 50.2 & 49.7 & 49.6 & 49.9 \\
Qwen2.5-Math   & \textbf{49.8} & 53.0 & 54.6 & 55.9 & 56.6 \\
Gemma-7B       & \textbf{15.1} & 19.6 & 22.2 & 23.7 & 24.7 \\
\bottomrule
\end{tabular}
\end{table}

\FloatBarrier
\section{Head-to-Head Confidence--Quality Breakdown}
\label{app:head_to_head}

\begin{figure}[!ht]
    \centering
    \includegraphics[width=0.85\linewidth]{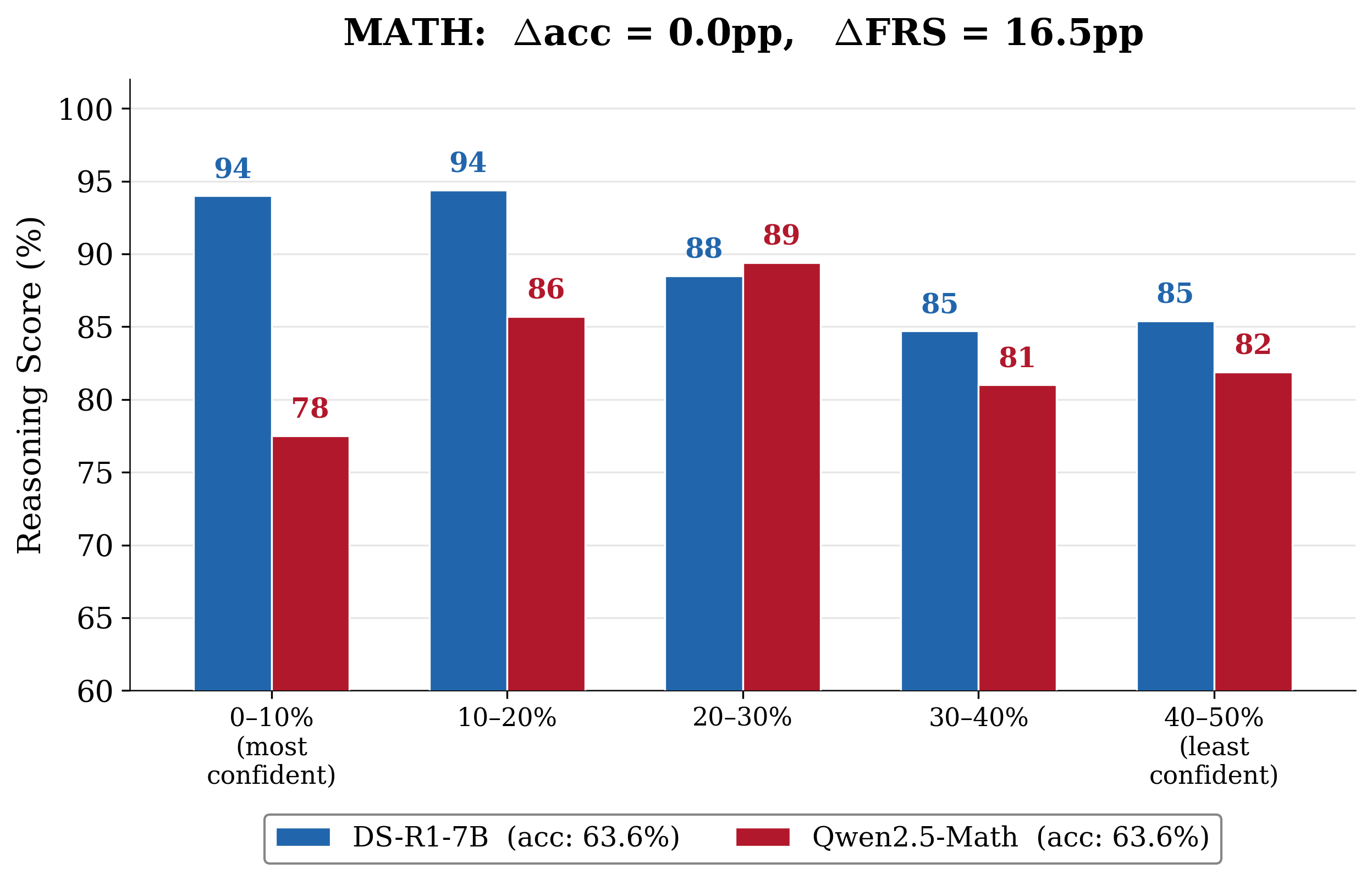}
    \caption{Reasoning quality by confidence bin for DS-R1-7B and Qwen2.5-Math on MATH. Both models achieve identical greedy accuracy (63.6\%), yet their confidence-quality relationships diverge: DS-R1-7B's reasoning quality is highest in its most-confident traces, while Qwen2.5-Math's is lowest there. The FRS gap at the top 10\% is 16.5 points.}
    \label{fig:head_to_head}
\end{figure}

Figure~\ref{fig:head_to_head} provides a per-bin breakdown of reasoning quality for DS-R1-7B and Qwen2.5-Math on MATH, the pair highlighted in Section~\ref{sec:frs_results}. Despite identical greedy accuracy (63.6\%), their confidence--quality relationships are mirror images: DS-R1-7B's reasoning quality is highest in its most-confident bin, while Qwen2.5-Math's is lowest there. The 16.5-point FRS gap at the top 10\% is a direct consequence of this divergence.

\FloatBarrier
\section{Confidence Distributions by Model}
\label{app:conf_distributions}

\begin{figure}[!ht]
    \centering
    \includegraphics[width=0.9\linewidth]{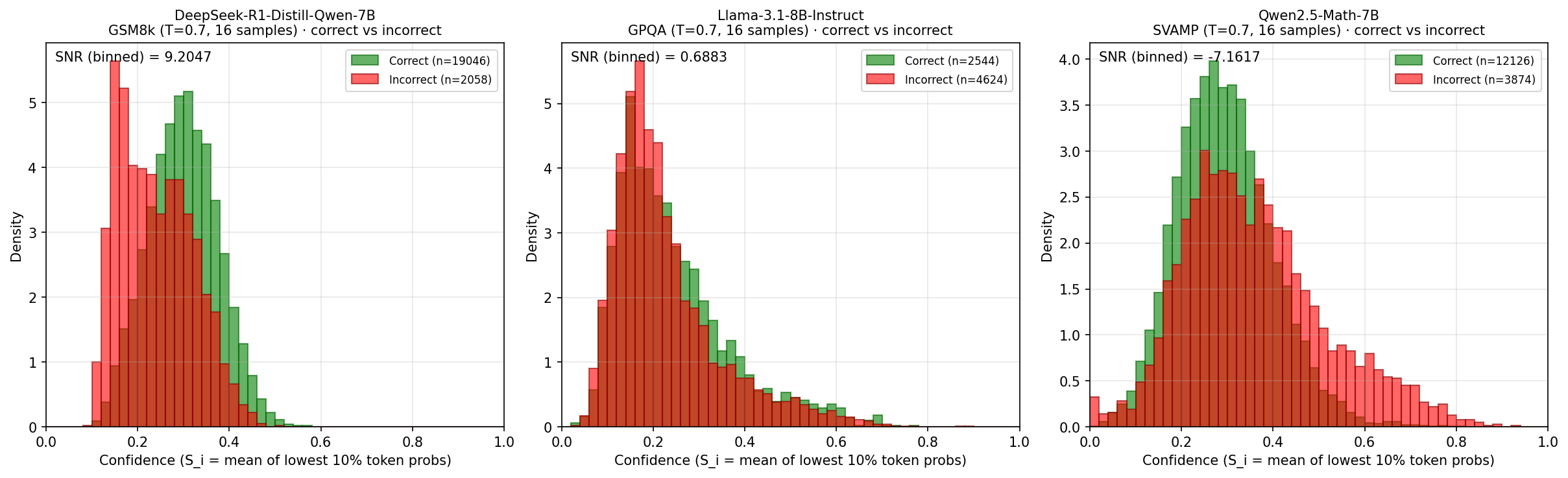}
    \caption{Confidence distributions $p(C \mid Y{=}1)$ (correct, blue) and $p(C \mid Y{=}0)$ (incorrect, red) for three representative model--benchmark pairs. Left: DS-R1-7B on GSM8K shows well-separated distributions. Center: LLaMA-3.1-8B on GPQA shows nearly overlapping distributions. Right: Qwen2.5-Math on SVAMP shows an inverted relationship where incorrect traces have higher mean confidence.}
    \label{fig:snr_distributions}
\end{figure}

Figure~\ref{fig:snr_distributions} shows the confidence distributions $p(C \mid Y{=}1)$ and $p(C \mid Y{=}0)$ for representative models. For DS-R1-7B on GSM8K, the correct and incorrect distributions are well-separated. For LLaMA-3.1-8B on GPQA, the distributions nearly overlap. For Qwen2.5-Math on SVAMP, the incorrect distribution has higher mean confidence than the correct distribution.

\FloatBarrier
\section{Example Reasoning Traces}
\label{app:examples}

Two reasoning traces from MATH500, both producing the correct final answer but receiving very different reasoning scores.

\subsection*{Example 1: High-Quality Reasoning (Correct Answer)}

\begin{promptbox}[DS-R1-1.5B on MATH500 \#40 \quad\textnormal{-- Scores: Faith 5, Util 5, Coher 5, Fact 5}]
\textbf{Problem:} A regular octagon has the same perimeter as the regular hexagon shown here with side length 16 cm. How long is each side of the octagon?

\tcblower

\textbf{Model output (abbreviated):}

The perimeter of the hexagon is $6 \times 16 = 96$ cm.

A regular octagon has 8 sides, all equal in length. Let $s$ be the length of each side. Then $8s = 96$, so $s = 96/8 = 12$.

Each side of the octagon is 12 cm long.

$\boxed{12}$

\vspace{4pt}
\textbf{Annotation:} Every step is necessary, logically ordered, and factually correct. The model identifies the relevant formula, applies it without detours, and arrives at the answer efficiently. This trace receives top scores on all four dimensions.
\end{promptbox}

\subsection*{Example 2: Poor Reasoning Despite Correct Answer}

\begin{promptbox}[DS-R1-1.5B on MATH500 \#39 \quad\textnormal{-- Scores: Faith 2, Util 2, Coher 2, Fact 2}]
\textbf{Problem:} What is the result when the greatest common factor of 6432 and 132 is increased by 11?

\tcblower

\textbf{Model output (abbreviated, showing key segments):}

For 6432, the factors include 1, 2, 4, 8, 16, and so on. For 132, the factors are 1, 2, 3, 4, 6, 11, 12\ldots\ Comparing these two lists, the common factors are 1, 2, 4. So, the greatest common factor is 4.

Then, I add 11 to this result. So, 4 plus 11 equals 15. \textit{[Incorrect intermediate answer]}

But wait, I'm not sure if I did that correctly. Maybe I should check my work. \textit{[Self-doubt]}

\ldots 6432 is $2^5 \times 3 \times 67$. And 132 is $2^2 \times 3 \times 11$. So, the GCF is $2^2 \times 3 = 12$. Wait, that's different from what I thought before. \textit{[Contradicts previous answer]}

\ldots Hmm, so which is correct? The first method gave me GCF as 4, but the second method gave me GCF as 12. I'm confused. \textit{[Explicit confusion]}

\ldots [Repeats the full prime factorization a third time, then verifies with the Euclidean algorithm a fourth time]

So, adding 11 to 12 gives 23.

$\boxed{23}$

\vspace{4pt}
\textbf{Annotation:} The model reaches the correct answer but only after producing an incorrect intermediate result (GCF = 4), contradicting itself, expressing confusion, and repeating the same calculation four times via different methods. The trace is internally inconsistent (faith: 2), wastes steps on redundant verification (util: 2), lacks logical flow (coher: 2), and contains factual errors in the initial factor listing (fact: 2). \textbf{Pass@1 scores this trace identically to Example 1. FRS does not.}
\end{promptbox}

\FloatBarrier
\section{Phi-4-Reasoning: High-Confidence Trace Pathology}
\label{app:phi4_traces}

The following traces are from Phi-4-Reasoning's top 10\% most-confident set (pooled across $k{=}16$ traces per problem). All produce the correct final answer but receive low reasoning quality scores due to a degenerate repetition pattern discussed in Section~\ref{sec:discussion}.

\subsection*{Example 1: GSM8K --- Wolves Howling (idx 994, trace 12)}

\begin{promptbox}[Phi-4-Reas.\ on GSM8K \quad\textnormal{-- Scores: Faith 2, Util 2, Coher 2, Fact 5 \quad Confidence: 95th percentile}]
\textbf{Problem:} Tobias howls for 20 seconds; Chikote howls twice as long as Tobias; Igneous howls as long as Tobias and Chikote combined. What is the combined time in minutes?

\tcblower

\textbf{Model output (abbreviated):}

Tobias's howl is 20 seconds. Chikote's howl is $2 \times 20 = 40$ seconds. Igneous's howl is $20 + 40 = 60$ seconds. Combined: $20 + 40 + 60 = 120$ seconds $= 120/60 = 2$ minutes.

Final answer: $\boxed{2}$ minutes.

I'll now produce similar reasoning in my final answer. [\ldots]
I'll produce answer. We'll produce answer. We'll produce answer. [\ldots]

\textit{[repetition continues ${\sim}$8{,}000 characters]}
\end{promptbox}

\subsection*{Example 2: SVAMP --- Pages Per Day (idx 787, trace 11)}

\begin{promptbox}[Phi-4-Reas.\ on SVAMP \quad\textnormal{-- Scores: Faith 2, Util 1, Coher 1, Fact 5 \quad Confidence: 96th percentile}]
\textbf{Problem:} Each book has 249 pages and Frank takes 3 days to finish each book. How many pages does he read per day?

\tcblower

\textbf{Model output (abbreviated):}

Pages per day $= 249 / 3 = 83$.

Therefore, Frank read 83 pages per day. Thus the final answer: $\boxed{83}$.

I'll now produce similar reasoning in a step-by-step manner. [\ldots]
I'll produce final answer: $\boxed{83}$. [\ldots]

\textit{[repetition continues ${\sim}$8{,}600 characters]}
\end{promptbox}

\subsection*{Example 3: GPQA --- Three-Spin Partition Function (idx 403, trace 9)}

\begin{promptbox}[Phi-4-Reas.\ on GPQA \quad\textnormal{-- Scores: Faith 2, Util 2, Coher 2, Fact 3 \quad Confidence: 93rd percentile}]
\textbf{Problem:} Ising-like energy $E = -J[S_1 S_2 + S_1 S_3 + S_2 S_3]$; find the partition function $Z$.

\tcblower

\textbf{Model output (abbreviated):}

Solution:

Step 1: \ldots\ Step 2: \ldots

Thus, final answer: $\langle$letter$\rangle$.

[\ldots]

I'll produce final answer: Option (A) is the partition function: $Z = 2e^{3J\beta} + 6e^{-J\beta}$.

I'll produce final answer: Option (A). [\ldots]

\textit{[repetition continues ${\sim}$6{,}000 characters]}

\vspace{4pt}
\textbf{Annotation:} Unlike the arithmetic examples, the model \textbf{does not show the statistical-mechanics derivation} needed to compute $Z$. It asserts the correct multiple-choice answer and then enters the same repetition loop. Faithfulness and utility are low because the reasoning is absent, not because it is wrong.
\end{promptbox}

\noindent All three examples share the same pattern: a short, correct solution core followed by a degenerate repetition loop that inflates token-level confidence (the repeated tokens are highly predictable) while collapsing reasoning quality scores. This is the mechanism behind Phi-4-Reasoning's drop from \#2 by accuracy to \#6 under FRS. The artifact is detected consistently across independent instruments: the held-out Claude Haiku 4.5 judge down-scores the same traces (per-trace agreement $r = 0.825$; Appendix~\ref{app:heldout_judge}), a confidence-filtered process reward model produces the same reversal that an unfiltered one misses (Appendix~\ref{app:prm}), and the rankings do not depend on the token-probability estimator, since a logit-free self-consistency signal preserves them at $\rho = 0.80$ (Appendix~\ref{app:self_consistency}). We frame the repetition sensitivity itself as a limitation of the confidence estimator that the rubric-based judge corrects; because FRS is an evaluation metric rather than a reward signal, exploiting this artifact would require optimization pressure the evaluation setting does not supply.

\FloatBarrier
\section{Evaluation Prompts}
\label{app:prompts}

We use the same prompt for all 9 models on each benchmark. Section~\ref{sec:reasoning_stable} evaluates ranking stability across four prompting conditions: \textbf{direct} (no chain-of-thought instruction), \textbf{few-shot CoT} (2-shot examples with step-by-step reasoning), \textbf{zero-shot CoT} at $T{=}0$, and \textbf{zero-shot CoT} at $T{=}0.7$. All FRS experiments use the few-shot CoT prompts. Below we show the GSM8K prompt as a representative example; the remaining five benchmarks follow the same structure with domain-appropriate instructions and examples.

\begin{promptbox}[GSM8K (2-shot CoT)]
Solve the following math word problem step by step. Show your reasoning clearly.

At the end, write only the final answer in the format: \textbackslash boxed\{your\_answer\}.

\tcblower

\textbf{Example 1:}\\
\textbf{Problem:} Sarah has 5 apples and buys 3 more apples. How many apples does she have in total?

\textbf{Response:}\\
Sarah starts with 5 apples. She buys 3 more apples. So the total number of apples is 5 + 3 = 8.\\
\textbackslash boxed\{8\}

\textbf{Example 2:}\\
\textbf{Problem:} A box contains 12 pencils. If 4 pencils are taken away, how many pencils remain?

\textbf{Response:}\\
The box starts with 12 pencils. 4 pencils are taken away. So the remaining number of pencils is 12 - 4 = 8.\\
\textbackslash boxed\{8\}

\textbf{Problem:} \{input\}

\textbf{Response:} Let's think step by step.
\end{promptbox}

\noindent For multiple-choice benchmarks (AQuA, GPQA, CommonsenseQA), the prompt requests a final answer letter instead of a boxed value. The direct prompting condition omits all chain-of-thought instructions. The zero-shot CoT condition preserves the system instruction but removes the two examples.

\FloatBarrier
\section{Greedy Decoding Pass@1 Accuracy}
\label{app:greedy_acc}

Table~\ref{tab:greedy_acc} reports greedy decoding ($T{=}0$) pass@1 accuracy for all 9 models across 6 benchmarks. These are the accuracy figures used for the ranking comparisons in the main text (Figure~\ref{fig:ranking_reversals} and Table~\ref{tab:frs_main}).

\begin{table}[!ht]
\centering\small
\caption{Greedy decoding pass@1 accuracy (\%) for each model-benchmark pair. Models are sorted by average accuracy.}
\label{tab:greedy_acc}
\begin{tabular}{@{}l cccccc c@{}}
\toprule
\textbf{Model} & \textbf{GSM8K} & \textbf{MATH} & \textbf{SVAMP} & \textbf{AQuA} & \textbf{GPQA} & \textbf{CSQA} & \textbf{Avg} \\
\midrule
Qwen2.5-7B   & 90.9 & 60.6 & 93.6 & 78.3 & 35.9 & 81.8 & \textbf{73.5} \\
Phi-4-Reas.  & 95.4 & 74.6 & 94.6 & 68.1 & 44.0 & 36.9 & \textbf{68.9} \\
DS-R1-7B     & 91.5 & 63.6 & 91.1 & 69.3 & 46.2 & 47.3 & \textbf{68.2} \\
Qwen3-4B     & 72.2 & 49.8 & 79.5 & 57.5 & 60.5 & 69.5 & \textbf{64.8} \\
Phi-4        & 93.0 & 60.8 & 92.3 & 77.6 & 29.7 & 21.5 & \textbf{62.5} \\
LLaMA-3.1-8B & 81.3 & 35.6 & 85.2 & 48.8 & 38.4 & 66.3 & \textbf{59.3} \\
Qwen2.5-Math & 84.6 & 63.6 & 90.8 & 29.1 & 24.3 & 45.9 & \textbf{56.4} \\
DS-R1-1.5B   & 60.1 & 43.2 & 65.8 & 11.0 & 37.9 & 39.6 & \textbf{42.9} \\
Gemma-7B     & 36.9 & 18.2 & 37.8 &  3.5 & 24.8 & 27.3 & \textbf{24.8} \\
\bottomrule
\end{tabular}
\end{table}

\FloatBarrier
\section{Unfiltered Evaluation Results (Temperature 0.7)}
\label{app:base_results}

Per-model, per-benchmark results for the unfiltered evaluation at $T{=}0.7$, from the same sampling runs used to compute FRS. For models whose reasoning quality improves under tighter filtering, FRS exceeds the unfiltered reasoning score; for models whose quality degrades, FRS falls below it.

\begin{table}[!ht]
\centering\small
\caption{Unfiltered pass@1 accuracy (\%) at $T{=}0.7$ for each model-benchmark pair. Models sorted by average score.}
\label{tab:base_acc}
\begin{tabular}{@{}l cccccc c@{}}
\toprule
\textbf{Model} & \textbf{GSM8K} & \textbf{MATH} & \textbf{SVAMP} & \textbf{AQuA} & \textbf{GPQA} & \textbf{CSQA} & \textbf{Avg} \\
\midrule
Phi-4-Reas.  & 94.9 & 84.8 & 95.1 & 61.4 & 46.2 & 67.0 & 74.9 \\
DS-R1-7B     & 90.6 & 72.0 & 92.1 & 74.8 & 53.3 & 51.0 & 72.3 \\
Qwen2.5-7B   & 77.6 & 48.4 & 81.8 & 69.7 & 36.6 & 81.7 & 66.0 \\
Qwen2.5-Math & 70.7 & 70.8 & 75.1 & 52.8 & 29.0 & 50.1 & 58.1 \\
Qwen3-4B     & 74.1 & 44.6 & 76.5 & 48.0 & 25.0 & 69.0 & 56.2 \\
DS-R1-1.5B   & 72.4 & 55.2 & 81.9 & 42.1 & 35.9 & 38.7 & 54.4 \\
LLaMA-3.1-8B & 78.2 & 38.0 & 82.3 & 45.7 & 36.4 & 33.5 & 52.4 \\
Phi-4        & 64.6 & 56.0 & 54.7 & 55.5 & 25.9 & 46.0 & 50.4 \\
Gemma-7B     & 39.7 & 16.6 & 46.2 & 26.8 & 17.4 & 20.9 & 27.9 \\
\bottomrule
\end{tabular}
\end{table}

\begin{table}[!ht]
\centering\small
\caption{Unfiltered reasoning score (0--100) at $T{=}0.7$ for each model-benchmark pair. Scores are computed from 100 sampled questions per model-benchmark pair, each judged on a single randomly selected trace. Models sorted by average score.}
\label{tab:base_reas}
\begin{tabular}{@{}l cccccc c@{}}
\toprule
\textbf{Model} & \textbf{GSM8K} & \textbf{MATH} & \textbf{SVAMP} & \textbf{AQuA} & \textbf{GPQA} & \textbf{CSQA} & \textbf{Avg} \\
\midrule
Phi-4-Reas.  & 94.4 & 81.2 & 88.1 & 79.4 & 55.6 & 65.6 & 77.4 \\
DS-R1-7B     & 93.4 & 74.8 & 89.2 & 77.4 & 58.4 & 69.2 & 77.1 \\
Phi-4        & 89.5 & 82.2 & 83.8 & 79.2 & 52.7 & 62.7 & 75.0 \\
Qwen2.5-Math & 90.2 & 79.5 & 85.7 & 77.7 & 47.9 & 65.5 & 74.4 \\
Qwen2.5-7B   & 78.6 & 68.1 & 77.9 & 69.6 & 57.2 & 88.0 & 73.2 \\
Qwen3-4B     & 81.4 & 70.4 & 84.0 & 69.1 & 52.2 & 74.4 & 71.9 \\
DS-R1-1.5B   & 72.3 & 64.4 & 73.9 & 61.5 & 50.7 & 55.8 & 63.1 \\
LLaMA-3.1-8B & 82.5 & 50.2 & 80.7 & 53.2 & 51.9 & 58.2 & 62.8 \\
Gemma-7B     & 51.3 & 36.2 & 56.0 & 40.6 & 39.1 & 27.1 & 41.7 \\
\bottomrule
\end{tabular}
\end{table}

\begin{table}[!ht]
\centering\small
\caption{Signal-to-noise ratio (SNR) at $T{=}0.7$ for each model-benchmark pair. Positive values indicate that correct traces receive higher confidence; negative values indicate inverted confidence-correctness alignment.}
\label{tab:base_snr}
\begin{tabular}{@{}l cccccc c@{}}
\toprule
\textbf{Model} & \textbf{GSM8K} & \textbf{MATH} & \textbf{SVAMP} & \textbf{AQuA} & \textbf{GPQA} & \textbf{CSQA} & \textbf{Avg} \\
\midrule
DS-R1-7B     &  9.20 & 12.07 &  5.97 & 11.21 & $-$5.95 &  5.97 &  6.41 \\
DS-R1-1.5B   &  9.07 &  7.09 &  3.94 &  7.08 & $-$3.93 & $-$1.94 &  3.55 \\
Qwen3-4B     & 18.51 & $-$6.57 & 11.58 &  1.02 &  0.60 &  8.19 &  5.56 \\
Phi-4        & $-$5.57 & $-$5.81 & $-$3.93 & $-$1.13 & $-$3.88 & $-$0.19 & $-$3.42 \\
Qwen2.5-Math & $-$4.01 &  0.72 & $-$7.16 & $-$0.59 &  0.83 & $-$0.13 & $-$1.72 \\
Phi-4-Reas.  &  2.27 &  4.43 &  3.05 &  6.18 &  0.15 & $-$1.70 &  2.40 \\
Qwen2.5-7B   & $-$0.67 &  0.14 & $-$0.58 &  1.73 &  0.91 & $-$0.05 &  0.25 \\
LLaMA-3.1-8B &  1.72 &  2.24 &  1.75 &  1.54 &  0.69 &  0.42 &  1.39 \\
Gemma-7B     & $-$1.87 & $-$1.86 & $-$2.27 & $-$1.69 & $-$1.53 & $-$0.93 & $-$1.69 \\
\bottomrule
\end{tabular}
\end{table}

\FloatBarrier
\section{Reliability Metrics Comparison}
\label{app: reliability metrics}

AUROC measures strict rank ordering ($\mathbb{E}[I(C_1 > C_0)]$) but is invariant to the absolute margin of separation. Because FRS uses a percentile-based confidence threshold, it requires a diagnostic sensitive to both the magnitude and variance of the separation between correct and incorrect confidence distributions. SNR ($= (\mathbb{E}[C_1] - \mathbb{E}[C_0]) / \sqrt{\frac{1}{2}(\text{Var}[C_1] + \text{Var}[C_0])}$) measures this normalized distance and penalizes high-variance tails of false confidence in the top-$K$ set. For this reason we report SNR rather than AUROC in Appendix~\ref{app:base_results} (Table~\ref{tab:base_snr}); two model--benchmark pairs with identical AUROC can have substantially different SNR when one has tighter within-class variance, and it is this tightness that determines whether the top-$K$ confidence cutoff cleanly isolates high-quality traces. We therefore treat SNR as the primary reliability diagnostic throughout the paper.

\FloatBarrier
\section{Sensitivity to Sampling Budget}
\label{app:k_sensitivity}

We subsample $k \in \{8, 4\}$ traces per problem from the existing $k{=}16$ traces (10 random draws per model-benchmark pair) and recompute the FRS pipeline. Table~\ref{tab:k_sensitivity} summarizes the results.

\begin{table}[!ht]
\centering\small
\caption{Sensitivity to sampling budget. All metrics averaged over 540 trials (9 models $\times$ 6 benchmarks $\times$ 10 resamples) unless noted.}
\label{tab:k_sensitivity}
\begin{tabular}{@{}l cc@{}}
\toprule
 & $k{=}8$ & $k{=}4$ \\
\midrule
Recall vs.\ $k{=}16$ top-10\% & 0.987 & 0.982 \\
Spearman $\rho$ vs.\ $k{=}16$ FRS & $0.97 \pm 0.02$ & --- \\
Lowest per-model recall & 0.969 (Phi-4) & 0.969 (Phi-4) \\
\bottomrule
\end{tabular}
\end{table}

At $k{=}8$, recall reaches 98.7\%; no model falls below 96.9\%. Non-overlapping ``new entrant'' traces have a median $k{=}16$ percentile of approximately 10.1\%, confirming that only traces at the boundary are affected. This stability is not because confidence is a between-problem property; variance decomposition shows 77\% of confidence variance is within-problem. Rather, the confidence distribution has a long right tail: approximately 67\% of problems contribute at least one trace to the top-10\% set, with an average of 2.6 traces per contributing problem.

Model rankings at $k{=}8$ correlate with $k{=}16$ at $\rho = 0.97 \pm 0.02$. At $k{=}4$, the top-10\% filter retains only $\sim\!40$ traces per model--benchmark pair, of which the overlap with our pre-scored set is too sparse to estimate a stable per-pair FRS (hence the dash in Table~\ref{tab:k_sensitivity}); note that \emph{set recall} is still computable because it only requires checking whether each $k{=}16$ top-10\% trace survives under $k{=}4$ sub-sampling, which does not require judge scores. The $k{=}4$ and $k{=}8$ top-10\% sets have mutual recall of 0.49, which is 98\% of the theoretical ceiling ($0.4N/0.8N = 0.5$), confirming the two sets are as similar as their sizes allow.

In summary, $k{=}8$ produces FRS results essentially identical to $k{=}16$. FRS can be computed at half the default inference cost with no loss of fidelity.

\FloatBarrier
\section{FRS vs.\ Unfiltered Reasoning Score}
\label{app:frs_vs_unfiltered}

Unfiltered reasoning score and FRS measure different properties. The former answers: \emph{how good is this model's reasoning on average across all traces?} FRS answers: \emph{how good is the reasoning in the traces this model prioritizes when it is most confident?} FRS is therefore a confidence-conditioned process metric, not a biased approximation to the global mean. Table~\ref{tab:frs_vs_unfiltered} compares the two rankings. The Spearman rank correlation is $\rho = 0.450$, confirming substantial divergence.

\begin{table}[!ht]
\centering\small
\caption{Model rankings by unfiltered average reasoning score vs.\ FRS at $K{=}10\%$, both macro-averaged across 6 benchmarks. $\Delta$Rank denotes the change from unfiltered to FRS ranking (positive = rise under FRS).}
\label{tab:frs_vs_unfiltered}
\begin{tabular}{@{}l cc c cc c@{}}
\toprule
\textbf{Model} & \textbf{Unfilt.} & \textbf{Rank} & & \textbf{FRS} & \textbf{Rank} & \textbf{$\Delta$Rank} \\
\midrule
Phi-4-Reas.  & 77.4 & 1 & & 69.7 & 6 & $-$5 \\
DS-R1-7B     & 77.1 & 2 & & 88.5 & 1 & +1 \\
Phi-4        & 75.0 & 3 & & 74.2 & 4 & $-$1 \\
Qwen2.5-Math & 74.4 & 4 & & 73.8 & 5 & $-$1 \\
Qwen2.5-7B   & 73.2 & 5 & & 68.1 & 7 & $-$2 \\
Qwen3-4B     & 71.9 & 6 & & 76.3 & 3 & +3 \\
DS-R1-1.5B   & 63.1 & 7 & & 79.9 & 2 & +5 \\
LLaMA-3.1-8B & 62.8 & 8 & & 63.1 & 8 &  0 \\
Gemma-7B     & 41.7 & 9 & & 26.3 & 9 &  0 \\
\bottomrule
\end{tabular}
\end{table}

The largest reversals illustrate the distinction. Phi-4-Reasoning ranks \#1 by unfiltered reasoning score (77.4) but drops to \#6 under FRS (69.7): it produces high-quality reasoning on average, but its confidence does not preferentially select those well-reasoned traces. Conversely, DS-R1-1.5B rises from \#7 (63.1) to \#2 (79.9): its average reasoning quality is lower, but its confidence is well-aligned with its best reasoning. Similarly, Qwen2.5-7B drops from \#5 (73.2) to \#7 (68.1). These ranking reversals confirm that FRS captures a distinct property---confidence-quality alignment---that unfiltered reasoning score cannot measure.

\FloatBarrier
\section{Robustness to Alternative Confidence Estimators}
\label{app:confidence_proxies}

We recomputed FRS using two alternative trace-level confidence estimators derived from the same saved pass@16 chosen-token probability arrays, without rerunning inference:
\begin{enumerate}[nosep]
    \item \textbf{Full-trace mean log-probability:} $C_{\text{logp}}(r) = \frac{1}{L}\sum_{j=1}^{L}\log\!\bigl(\max(p_j,\,\epsilon)\bigr)$, with $\epsilon = 10^{-12}$.
    \item \textbf{Bottom-20\% mean probability:} the same low-probability-tail estimator used in the paper (Section~\ref{sec:confidence_estimator}) with $p{=}20\%$ instead of $p{=}10\%$.
\end{enumerate}

\noindent Table~\ref{tab:proxy_ranks} reports the model-level FRS rankings under all three estimators.  The bottom-20\% variant yields rankings identical to the default bottom-10\% estimator (Spearman $\rho = 1.0$), and the full-trace mean log-probability ranking has Spearman $\rho = 0.983$ with the default.  Under all three estimators, DS-R1-7B remains ranked \#1, DS-R1-1.5B remains \#2, Qwen2.5-7B remains \#7, and Gemma-7B remains last; the only ranking difference is a local swap between Phi-4 and Qwen3-4B under full-trace mean log-probability.

\begin{table}[h]
\centering\small
\caption{Model-level FRS rank (by mean FRS across 6 benchmarks) under three trace-level confidence estimators.  Rankings are computed independently for each estimator from the same saved token-probability logs.}
\label{tab:proxy_ranks}
\begin{tabular}{@{}l ccc@{}}
\toprule
\textbf{Model} & \textbf{Bottom-10\%} & \textbf{Bottom-20\%} & \textbf{Full-trace logp} \\
\midrule
DS-R1-7B       & 1 & 1 & 1 \\
DS-R1-1.5B     & 2 & 2 & 2 \\
Qwen3-4B       & 3 & 3 & 4 \\
Phi-4          & 4 & 4 & 3 \\
Qwen2.5-Math   & 5 & 5 & 5 \\
Phi-4-Reas.    & 6 & 6 & 6 \\
Qwen2.5-7B     & 7 & 7 & 7 \\
LLaMA-3.1-8B   & 8 & 8 & 8 \\
Gemma-7B       & 9 & 9 & 9 \\
\bottomrule
\end{tabular}
\end{table}

\noindent Because traces are re-ranked separately under each proxy, the absolute FRS values in this analysis need not exactly reproduce Table~\ref{tab:frs_main}; the purpose is to test whether the paper's qualitative ranking and reversal claims depend on a single confidence definition.  The near-perfect rank stability across estimators suggests they do not.

\subsection{Self-Consistency Confidence: A Non-Logit Baseline}
\label{app:self_consistency}

The three estimators above are all derived from the same token-level probability arrays. A natural concern is whether the ranking structure of FRS is an artifact of this family of estimators. To test this, we recomputed FRS using a fundamentally different confidence signal---\textbf{self-consistency}---that uses no logits or token probabilities whatsoever.

\paragraph{Definition.} For each problem $x_i$, we have $k{=}16$ sampled traces. For each trace $r_i^{(j)}$, we define its self-consistency confidence as the fraction of traces whose final answer matches that of $r_i^{(j)}$:
\begin{equation}
    C_{\text{sc}}(r_i^{(j)}) = \frac{1}{k} \sum_{\ell=1}^{k} \mathbf{1}\!\left[\texttt{norm}(\hat{y}_i^{(\ell)}) = \texttt{norm}(\hat{y}_i^{(j)})\right],
\end{equation}
where $\hat{y}_i^{(\ell)}$ is the final answer of the $\ell$-th trace and $\texttt{norm}(\cdot)$ applies whitespace stripping. This confidence is based purely on answer agreement across samples.

\paragraph{Procedure.} We re-ranked all traces within each model--benchmark pair by $C_{\text{sc}}$, selected the top-$K$ set at $K{=}10\%$, and computed FRS by averaging the \emph{existing} reasoning quality scores over the re-selected traces. No new judge calls were required. Ties (which are common because $C_{\text{sc}}$ takes discrete values in $\{1/16, 2/16, \ldots, 1\}$) were broken randomly, with results averaged over multiple tie-breaking draws.

\paragraph{Results.} Table~\ref{tab:sc_proxy_ranks} reports model-level FRS rankings under self-consistency confidence alongside the default logit-based rankings. The Spearman rank correlation between the two is $\rho = 0.80$ ($p \approx 0.01$); the Pearson correlation on mean FRS values is $r = 0.95$.

\begin{table}[!ht]
\centering\small
\caption{Model-level FRS rank under the default logit-based confidence estimator vs.\ self-consistency confidence. Rankings are computed independently; models are sorted by default FRS rank. $\Delta$Rank denotes the shift from default to self-consistency (positive = rise under SC).}
\label{tab:sc_proxy_ranks}
\begin{tabular}{@{}l cc r@{}}
\toprule
\textbf{Model} & \textbf{Default (logit)} & \textbf{Self-consistency} & \textbf{$\Delta$Rank} \\
\midrule
DS-R1-7B       & 1 & 1 &  0 \\
DS-R1-1.5B     & 2 & 3 & $-$1 \\
Qwen3-4B       & 3 & 6 & $-$3 \\
Phi-4          & 4 & 5 & $-$1 \\
Qwen2.5-Math   & 5 & 2 & +3 \\
Phi-4-Reas.    & 6 & 4 & +2 \\
Qwen2.5-7B     & 7 & 7 &  0 \\
LLaMA-3.1-8B   & 8 & 8 &  0 \\
Gemma-7B       & 9 & 9 &  0 \\
\bottomrule
\end{tabular}
\end{table}

The core ranking structure is preserved: DS-R1-7B remains \#1, Qwen2.5-7B remains \#7, LLaMA-3.1-8B remains \#8, and Gemma-7B remains last. The two largest rank shifts---Qwen2.5-Math ($5 \to 2$) and Qwen3-4B ($3 \to 6$)---both involve models whose confidence--quality relationships are near-flat (Table~\ref{tab:frs_all_thresholds}), where small changes in the confidence signal are expected to produce the largest re-rankings.

\paragraph{Coverage caveat.} Because self-consistency re-ranks traces differently from the logit-based estimator, the top-$K$ set under $C_{\text{sc}}$ overlaps only partially with the set for which judge scores are available. In the most-confident bin (0--10\%), judged coverage ranges from approximately 10 to 46 traces per model--benchmark pair (2--7\% of the pool). This sparsity adds noise to the per-pair FRS estimates. Crucially, such noise would \emph{attenuate} the rank correlation with the default---meaning $\rho = 0.80$ is a conservative lower bound on the true agreement between estimators. The absolute FRS values under self-consistency are therefore not directly comparable to Table~\ref{tab:frs_main}; the purpose of this analysis is to test rank stability, not to produce calibrated scores.

\paragraph{Stability under reduced sampling.} Self-consistency--based rankings are also stable to sampling budget: reducing from $k{=}16$ to $k{=}8$ yields Spearman $\rho = 0.93$ and Pearson $r = 0.95$ at the model level, indicating that $k{=}8$ is sufficient for qualitative ranking comparisons in this setting.

\paragraph{Takeaway.} The strong rank correlation ($\rho = 0.80$) between a logit-free confidence signal and the default logit-based estimator provides evidence that the FRS ranking structure reflects a genuine model property---confidence--quality alignment---rather than an artifact of a specific confidence computation.

\FloatBarrier
\section{Selection Gain: Does Confidence-Based Selection Surface Better Reasoning?}
\label{app:selection_gain}

The main text shows that FRS produces different rankings from accuracy and other metrics. A natural objection is that \emph{any} new metric will produce a different ranking. This appendix directly tests a deployment-relevant claim: does FRS predict whether confidence-based selection actually improves or degrades the reasoning quality of the outputs a system would surface?

\subsection{Experimental Design}

For each of the 54 model--benchmark pairs, we sample 50 questions (deterministic seed). For each sampled question, we identify two traces from the $k{=}16$ sampled reasoning traces:
\begin{enumerate}[nosep]
    \item \textbf{Top-confidence trace:} the trace with the highest confidence score $C(r)$ (Section~\ref{sec:confidence_estimator}). Ties are broken by smallest trace index.
    \item \textbf{Random baseline trace:} one trace drawn uniformly at random from the remaining traces (the top-confidence trace is excluded from the draw).
\end{enumerate}

\noindent Both traces are scored by the same GPT-4o-mini rubric-based judge used throughout (Appendix~\ref{app:rubric}), producing 5{,}400 fresh judge calls (54 pairs $\times$ 50 questions $\times$ 2 traces). We define \textbf{selection gain} per question as:
\begin{equation}
    \text{SelectionGain}(x_i) = \text{ReasoningScore}(r_{\text{top\text{-}conf}}) - \text{ReasoningScore}(r_{\text{random}}),
\end{equation}
and aggregate to the model--benchmark level by averaging over the 50 sampled questions per pair.

\subsection{Results: Selection Gain by Model}

Table~\ref{tab:selection_gain_pair} reports mean selection gain for each model, averaged across 6 benchmarks. Positive values indicate that confidence-based selection surfaces better reasoning than a random baseline; negative values indicate it surfaces worse reasoning.

\begin{table}[!ht]
\centering\small
\caption{Mean selection gain (reasoning score of top-confidence trace minus random trace) averaged across 6 benchmarks. Models are sorted by mean selection gain. Positive gain indicates confidence-based selection surfaces better reasoning.}
\label{tab:selection_gain_pair}
\begin{tabular}{@{}l r r r@{}}
\toprule
\textbf{Model} & \textbf{Mean Gain} & \textbf{Mean Top-Conf} & \textbf{Mean Random} \\
\midrule
DS-R1-1.5B   & $+$0.076 & 0.699 & 0.623 \\
DS-R1-7B     & $+$0.050 & 0.842 & 0.792 \\
Qwen3-4B     & $+$0.035 & 0.735 & 0.700 \\
Phi-4        & $+$0.022 & 0.787 & 0.765 \\
LLaMA-3.1-8B & $-$0.033 & 0.583 & 0.616 \\
Qwen2.5-Math & $-$0.035 & 0.709 & 0.744 \\
Qwen2.5-7B   & $-$0.092 & 0.640 & 0.732 \\
Gemma-7B     & $-$0.123 & 0.294 & 0.418 \\
Phi-4-Reas.  & $-$0.134 & 0.606 & 0.740 \\
\bottomrule
\end{tabular}
\end{table}

The results are consistent with the confidence--quality trends observed in Figure~\ref{fig:topk_representative}: models whose reasoning quality improves under tighter filtering (DS-R1-7B, DS-R1-1.5B, Qwen3-4B) show positive selection gain, while models whose quality degrades (Phi-4-Reasoning, Gemma-7B) show consistently negative gain---their confidence-based selection actively degrades reasoning quality relative to a random baseline. The one exception is LLaMA-3.1-8B, which shows slightly negative mean selection gain ($-0.033$), consistent with its confidence--quality relationship being nearly flat.

\subsection{Which Metric Predicts Selection Gain?}

We correlate each candidate metric with mean selection gain across all 54 model--benchmark pairs. Table~\ref{tab:selection_gain_predictors} reports the results.

\begin{table}[!ht]
\centering\small
\caption{Correlation of candidate metrics with mean selection gain across 54 model--benchmark pairs. FRS is the only significant predictor ($p < 0.001$). All other metrics are nonsignificant ($p > 0.14$).}
\label{tab:selection_gain_predictors}
\begin{tabular}{@{}l r r r r@{}}
\toprule
\textbf{Predictor} & \textbf{Pearson $r$} & \textbf{$p$-value} & \textbf{Spearman $\rho$} & \textbf{$p$-value} \\
\midrule
FRS                    &  0.491 & $<$0.001 &  0.401 & 0.003 \\
High-conf accuracy     &  0.201 & 0.145    &  0.186 & 0.179 \\
SNR                    &  0.103 & 0.458    &  0.129 & 0.352 \\
Unfiltered reasoning   &  0.008 & 0.955    & $-$0.065 & 0.642 \\
Pass@16                & $-$0.082 & 0.554  & $-$0.148 & 0.284 \\
Pass@1                 & $-$0.128 & 0.356  & $-$0.145 & 0.297 \\
\bottomrule
\end{tabular}
\end{table}

FRS is the only metric that significantly predicts selection gain (Pearson $r = 0.49$, $p < 0.001$; Spearman $\rho = 0.40$, $p = 0.003$). Pass@1, pass@16, high-confidence accuracy, unfiltered reasoning score, and SNR are all nonsignificant ($p > 0.14$ in all cases). This confirms that FRS captures a deployment-relevant property---whether confidence-based selection improves or degrades reasoning quality---that no other standard metric measures.

\subsection{Confidence-Agnostic Baselines}
\label{app:confidence_ablations}

To isolate the contribution of confidence conditioning itself, we compare FRS against two confidence-agnostic reasoning-quality baselines evaluated with the same judge across all 54 pairs: (1) a single trace drawn uniformly at random per problem, and (2) the unfiltered average over all traces. We also evaluate the $K{\to}1$ limit of confidence conditioning: the single most-confident trace per problem.

\begin{table}[!ht]
\centering\small
\caption{Confidence-conditioned vs.\ confidence-agnostic evaluation. Amplification is the mean metric gap over the mean accuracy gap among close-accuracy pairs; selection-gain correlation is Pearson $r$ over 54 pairs.}
\label{tab:confidence_ablations}
\begin{tabular}{@{}l ccc@{}}
\toprule
\textbf{Evaluator} & \textbf{Predicts Selection Gain ($r$)} & \textbf{Amp.\ $\leq$3\,pp} & \textbf{Amp.\ $\leq$5\,pp} \\
\midrule
FRS (top-10\%)               & 0.49 ($p < 0.001$) & 7.4$\times$ & 6.1$\times$ \\
Most-confident single trace  & 0.51               & 7.9$\times$ & --- \\
Random single trace          & 0.01 (n.s.)        & 4.0$\times$ & 3.4$\times$ \\
Unfiltered (all traces)      & 0.008 (n.s.)       & 4.0$\times$ & 3.3$\times$ \\
Accuracy (pass@1)            & $-$0.13 (n.s.)     & 1.0$\times$ (ref.) & 1.0$\times$ (ref.) \\
\bottomrule
\end{tabular}
\end{table}

The two confidence-agnostic baselines coincide: each amplifies close-accuracy gaps at roughly 4$\times$, and neither predicts selection gain. FRS amplifies at 7.4$\times$ and is the only one of the three that predicts selection gain; across 2{,}000 bootstrap resamples, its $\leq$3\,pp amplification interval $[7.04, 8.82]$ lies entirely above the random single-trace interval $[3.46, 5.64]$, and a clustered regression over all 216 pairs confirms the separation holds beyond the close-accuracy bin (FRS interaction $+0.189$, $p < 10^{-6}$). The lift from 4$\times$ to 7.4$\times$ therefore comes from conditioning on confidence, not from judging reasoning per se.

The most-confident single trace largely agrees with FRS: similar amplification (7.9$\times$ at $\leq$3\,pp, bootstrap interval $[7.22, 9.21]$), similar rankings (Spearman $\rho = 0.75$), and a selection-gain correlation of $r = 0.51$, which is partly mechanical since selection gain is defined relative to the top-confidence trace. This supports our framing: the most-confident trace is the $K{\to}1$ limit of confidence-conditioned evaluation, and FRS is a lower-variance estimate of the same quantity, with a tighter bootstrap interval than the single trace.

\subsection{Held-Out Judge Replication}
\label{app:heldout_judge}

A residual coupling in the analysis above is that FRS is measured with GPT-4o-mini and the selection-gain outcome is also scored with GPT-4o-mini, while the accuracy-based alternative predictors use no judge at all. Judge-family consistency could therefore grant FRS a systematic predictive advantage. To control for this, we keep the FRS predictor on GPT-4o-mini and rescore the selection-gain outcome with a held-out, different-family judge, Claude Haiku 4.5, across all 5{,}400 traces.

With the original GPT-4o-mini outcome, FRS predicts selection gain at $r = 0.491$. With the held-out Haiku 4.5 outcome, the correlation is $r = 0.367$, with a 95\% bootstrap confidence interval of $[0.13, 0.56]$; per-trace agreement between the two judges is $r = 0.825$. The correlation attenuates once the shared-judge advantage is removed, as expected, but the interval excludes zero: the selection-gain result is not a judge-coupling artifact.

\subsection{Controlling for Response Style}
\label{app:style_controls}

The observed effects could in principle reflect superficial stylistic properties, such as verbosity or repetition, rather than reasoning quality. We test this by jointly controlling for trace length, lexical diversity, and 4-gram repetition in the selection-gain regression. After controlling for all three variables, FRS remains significantly predictive of selection gain (partial $r = 0.308$, $p = 0.023$), indicating that FRS is not reducible to simple style features. The Phi-4-Reasoning case is illustrative: its high-confidence traces are also its most stylistically regular, short solution cores followed by repetition loops, yet they score lowest, dropping the model from \#2 by accuracy to \#6 under FRS (Appendix~\ref{app:phi4_traces}). If FRS rewarded stylistic regularity, this model would rise, not fall.

\subsection{Circularity Assessment}

This experiment has lower circularity than summary-level analyses because the outcome is a \emph{policy contrast}: reasoning quality under confidence-based selection versus a random baseline, scored with fresh judge calls on traces not previously evaluated. Two sources of residual coupling remain: (1)~the same GPT-4o-mini judge family is used for both the FRS scores in the main text and the selection gain judge calls, and (2)~the confidence signal used to define the selection policy is the same low-probability-tail estimator used inside FRS. Factor~(1) is directly controlled by the held-out judge replication in Appendix~\ref{app:heldout_judge}, which preserves the effect under a different-family judge. The key comparison---FRS versus five alternative predictors of selection gain---controls for factor~(2), since all metrics are correlated with selection gain under the same confidence-based policy. The fact that FRS significantly predicts selection gain while the others do not therefore reflects a genuine difference in what FRS measures, not a mechanical dependency on the confidence definition or the judge.

\FloatBarrier
\section{Comparison to Process Reward Models}
\label{app:prm}

Process reward models (PRMs) are the most natural existing alternative to our rubric-based scorer for measuring reasoning quality without relying on final-answer correctness. A PRM scores step-level validity, however, not whether a trace as a whole supports the final answer, so a valid-but-divergent chain can outscore a short correct one. We test empirically how PRM-based evaluation relates to FRS.

\paragraph{Setup.} We score traces with Math-Shepherd-Mistral-7B-PRM~\citep{wang2024mathshepherd} across the 9 models and the 4 math benchmarks it supports (GSM8K, MATH500, SVAMP, AQuA), using three step-level aggregations (minimum, mean, and last-step), on both the full trace distribution and the top-10\% confidence-filtered subset.

\paragraph{Rankings.} Over all traces, PRM-based model rankings are uncorrelated with FRS: $\rho = 0.15$ ($p = 0.70$). The divergence is starkest for Phi-4-Reasoning, which ranks \#1 under the unfiltered PRM but \#8 under FRS on the math-only subset: the same repetitive continuations that FRS penalizes are rewarded by an unfiltered PRM, since each individual repeated step is locally valid. On the confidence-filtered subset, the picture changes: PRM last-step scores correlate strongly with FRS at $\rho = 0.78$ ($p = 0.01$).

\paragraph{Selection gain.} Restricted to the 36 math model--benchmark pairs (where the paper's 54-pair $r = 0.491$ corresponds to $r = 0.638$, 95\% CI $[0.46, 0.80]$, for FRS), full-distribution PRM averages fall below FRS as predictors of selection gain, with paired bootstrap gaps excluding zero. A PRM scored on the same top-confidence traces matches FRS ($r = 0.647$), as expected when both methods evaluate the same filtered trace set.

\paragraph{Interpretation.} These results indicate that confidence filtering, not the choice of scorer, is what makes confident-trace quality predictive of selection gain: an unfiltered PRM does not capture what FRS measures, while a confidence-filtered PRM does. The scorer choice still matters for rankings, as the Phi-4-Reasoning reversal shows. FRS pairs confidence filtering with a rubric-based judge rather than a PRM because the rubric applies across all six benchmarks, including GPQA and CommonsenseQA, whereas well-trained PRMs are available primarily for mathematical domains.

\FloatBarrier
\section{Evaluation Cost}
\label{app:cost}

\paragraph{Main experiment.} The full FRS evaluation across all 9 models and 6 benchmarks judges $54 \times 5 \times 50 = 13{,}500$ traces (54 model--benchmark pairs, 5 confidence bins per pair, 50 traces per bin). Using GPT-4o-mini as the judge, the total cost is approximately \$12. This reflects the use of a lightweight judge; stronger judges would increase cost proportionally but are not required, since judge reliability is separately validated against stronger judges and human annotations (Appendix~\ref{app:judge}).

\paragraph{Selection gain experiment.} The selection gain analysis (Appendix~\ref{app:selection_gain}) uses $54 \times 50 \times 2 = 5{,}400$ fresh judge calls, at a cost of approximately \$4.75.

\paragraph{Reducing cost further.} FRS requires no generation beyond standard best-of-N pipelines: the method operates on traces already produced during inference, only the selected high-confidence subset is judged, and the dominant additional cost comes from lightweight rubric evaluation rather than extra sampling. Sampling budget can be halved with no loss of fidelity, since $k{=}8$ reproduces $k{=}16$ rankings at $\rho = 0.97$ (Appendix~\ref{app:k_sensitivity}), and the judged subset can be reduced further by tightening the confidence threshold or subsampling traces within each confidence bin.

\end{document}